\RequirePackage{snapshot}
\documentclass[12pt]{article}

\usepackage{graphicx}
\usepackage{amsmath,amssymb,amsfonts,amsthm}
\usepackage{hyperref}
\hypersetup{backref,colorlinks=true,citecolor=blue,linkcolor=blue,urlcolor=blue}

\usepackage[round]{natbib}

\usepackage{array}

\usepackage[left=1.25in,top=1in,right=1.25in,bottom=1in,nohead]{geometry}
\usepackage{setspace}

\usepackage[lined,ruled]{algorithm2e}
\SetKw{KwSet}{Set}
\SetKw{KwIterate}{Iterate}
\SetKwInOut{Input}{input}\SetKwInOut{Output}{output}

\usepackage{bm}
\usepackage{color} 

\usepackage{aliascnt}
\newtheorem{theorem}{Theorem}[section]

\newaliascnt{lemma}{theorem}

\aliascntresetthe{lemma}

\newaliascnt{cor}{theorem}
\newtheorem{corollary}[cor]{Corollary}
\aliascntresetthe{cor}

\newaliascnt{def}{theorem}

\aliascntresetthe{def}

\newtheorem{remark}{Remark}[section]

\newtheorem{condition}{Condition}

\newcommand{\email}[1]{\href{mailto:#1}{#1}}

\definecolor{greenmain}{rgb}{.2,.5,.2}

\newcommand{\random}[1]{\textrm{\tt Random}_{#1}}
\newcommand{\band}{\texttt{Band}}

\renewcommand{\hat}{\widehat}
\newcommand{\N}{Nystr\"om }
\newcommand{\CS}{column-sampling }

\newcommand{\Ran}{\textrm{ran}}
\newcommand{\X}{\mathbf{X}}
\newcommand{\x}{\mathbf{x}}
\newcommand{\inner}{\mathbf{S}}
\renewcommand{\outer}{\mathbf{Q}}
\newcommand{\1}{\mathbf{1}}
\newcommand{\Xbar}{\overline{\X}}

\newcommand{\A}{\mathbf{A}}
\newcommand{\Aone}{\mathbf{A}^{(1)}}
\newcommand{\Atwo}{\mathbf{A}^{(2)}}

\renewcommand{\O}{\mathbf{O}}

\newcommand{\V}{\mathcal{V}}

\newcommand{\G}{\mathcal{G}}
\renewcommand{\H}{\mathcal{H}}

\renewcommand{\S}{\Phi}
\newcommand{\I}{\mathbf{I}}

\newcommand{\ran}{\mbox{\normalfont ran}}

\newcommand{\gap}[1]{\textrm{\normalfont gap}_{d}\!\left(#1\right)}

\newcommand{\R}{\mathbb{R}}

\newcommand{\rootlq}[2]{\kappa_{#1 #2}}
\newcommand{\trace}[1]{\textrm{\normalfont trace}\!\left(#1\right)}

\begin{document}

\title{On the Nystr\"om and Column-Sampling Methods for the Approximate Principal Components Analysis of Large Data Sets\thanks{Darren Homrighausen is
    Assistant Professor, Department of Statistics, Colorado State
    University, Fort Collins, CO
    80523 (\email{darrenho@stat.colostate.edu}); Daniel J. McDonald
    is Assistant Professor, Department of Statistics, Indiana
    University, Bloomington, IN 47408 (\email{dajmcdon@indiana.edu}).}}

\author{Darren Homrighausen\\Colorado State University\and Daniel
  J.\ McDonald\\Indiana University}

\date{Version: \today}


\maketitle

\begin{abstract} 
  In this paper we analyze approximate methods for undertaking a
  principal components analysis (PCA) on large data sets. PCA is a
  classical dimension reduction method that involves the projection of
  the data onto the subspace spanned by the leading
  eigenvectors of the covariance matrix. This projection can be used
  either for exploratory purposes or as an input for further analysis,
  e.g.\ regression. If the data have billions of entries or more, the
  computational and storage requirements for saving and manipulating the
  design matrix in fast memory is prohibitive. Recently, the \N and
  \CS methods have appeared in the
  numerical linear algebra community
  for the randomized approximation of the
  singular value decomposition of large matrices.  However, their utility for
  statistical applications remains unclear. We compare
  these approximations
  theoretically by bounding the distance between the induced subspaces and
  the desired, but computationally infeasible, PCA subspace.  Additionally
  we show empirically, through simulations and a real data example involving a corpus
  of emails, the trade-off of approximation accuracy and computational complexity.

  \vspace{1em}
\end{abstract}

\section{Introduction} 

In modern
statistical applications such as genomics, neural image analysis, or
text analysis, the number of covariates $p$ is often extremely large. 
In this large-$p$ regime, dimension
reduction becomes a necessity both for interpretion and for
accurate prediction.  Though there are numerous methods for
reducing the dimensionality of the data --- multidimensional
scaling, discriminant analysis, locally linear embeddings, Laplacian
eigenmaps, and many others --- the first, and perhaps the most widely
used method, is principal components analysis or PCA
\citep{jollife2002principal}.

Suppose our
data is $n$ observations, each containing the values of
$p$ measurements.  Notationally, we represent this as having
observations $\tilde{X}_1, \ldots, \tilde{X}_n$, where $\tilde{X}_i =
(\tilde{X}_{i1}, \ldots,\tilde{X}_{ip})^{\top} \in \mathbb{R}^p$.
After concatenating our data into the matrix $\tilde{\X} =
[\tilde{X}_1,\ldots,\tilde{X}_n]^{\top} \in \R^{n\times p}$, we form
the matrix $\X$ by centering the columns of $\tilde{\X}$ with the matrix
$\Xbar = n^{-1}\1\1^{\top} \tilde{\X}$; that is, $\X = \tilde{\X} - \Xbar$.
Write the (reduced) singular value decomposition (SVD) of $\X$ as
\begin{equation}
  \X = U(\X) \Lambda(\X) V(\X)^{\top} =: U \Lambda V^{\top}, 
  \label{eq:svdX}
\end{equation}
where, for any matrix $\A$, we define $U(\A)$, $V(\A)$, and 
\begin{equation}
  \Lambda(\A) = \textrm{diag}(\lambda_1(\A),
  \ldots, \lambda_r(\A))
  \label{eq:LambdaDef}
\end{equation} 
to be the left and right singular vectors and the singular values of
the matrix $\A$, respectively, and $r$ is the rank of $\A$.  To
simplify notation, we write the singular vectors and values of $\X$ as
just $U,V$ and $\Lambda$ and reserve the functional notation for
use with other matrices.   

For each $d \leq r$, PCA seeks to find a projection that minimizes 
the squared error distance between the data and the projected data (see e.g.
\cite{jollife2002principal} for details).
These projections are given by the first $d$ columns of $U$ and $V$.
For example, if using PCA for regression with response vector $Y$,
the fitted values can be written $\hat Y = U_d{U_d}^{\top}Y$.
Here, the notation $\A_d$ is the matrix comprised of the first
$d$ columns of $\A$ and for $\Lambda_d$ we implicitly assume that the
vector $(\lambda_1(\A), \ldots, \lambda_r(\A))$ gets truncated to
length $d$ before being transformed into a diagonal matrix as in
equation \eqref{eq:LambdaDef}.  


\textbf{A Big Problem.} For small and medium sized problems in data analysis, PCA
 provides a powerful method for reducing the
dimension of the data via the decomposition in 
equation \eqref{eq:svdX}.  However, 
in modern applications, practitioners are routinely faced with
data volumes that seemed unimaginable even a decade ago.  In 2000,
humans produced 800,000 petabytes of stored
data.  This number is expected to grow to 35 zettabytes ($3.5\times
10^{22}$ bytes) by 2020. The social media website Twitter.com alone produces 7
terabytes daily~\citep{zikopoulos2011understanding}.  In just four hours on ``black Friday''
2012, Walmart handled 10 million cash register transactions and sold nearly
5,000 items per second~\citep{Wal-Mart-Stores2012}.  Airport security software must handle an arbitrarily  large
database of high-quality facial images, where each image could have millions of
pixels. 
Storing and processing this data for
statistical analysis becomes infeasible even with ever
increasing computer technology. 
These are indications that a practicing, applied statistician should
expect to be confronted with very large data sets that need
analysis.


For very large problems, PCA encounters two major practical issues:
processing constraints and memory constraints.  The computational complexity
is dominated by the cost of computing
the SVD of $\X$, defined in equation \eqref{eq:svdX},
which, if $p > n$, is $O(p^2n + n^3)$.  If $n$ is rather small, then
this computation has quadratic complexity, which can be computationally feasible.
However, if $n$ is also very large, say $n \approx p$, then the complexity is
$O(n^3)$, which is infeasible.  In addition
to the computational cost, there is an irreducible space cost to
storing dense matrices in fast memory.  

%



\subsection{Approximation Methods}
\label{sec:approximationMethods}



Suppose that $\A \in \mathbb{R}^{q\times q}$ is a symmetric,
nonnegative definite matrix with rank $r$; that is, for all $a \in \mathbb{R}^{q}$,
$a^{\top} \A a \geq 0$ and $\A^{\top} = \A$. To approximate 
$\A$, we fix an integer $l \ll q$ and form a sketching matrix $\S \in
\R^{q \times l}$.  Then, we report the following approximation: $\A
\approx (\A\S)(\S^{\top}\A\S)^{\dagger}(\A\S)^{\top}$.  Here, we define
$\A^{\dagger} := V(\A) \Lambda(\A)^{\dagger} V(\A)^{\top}$ to be the
Moore-Penrose pseudo inverse of $\A$ with  
$\Lambda(\A)^{\dagger} := \textrm{diag}(\lambda_1(\A)^{-1},\ldots,
\lambda_r(\A)^{-1},0,\ldots,0) \in \R^{q\times q}$.

The details behind the formation of the matrix $\S$ control the type of
approximation.
For the \N and \CS methods, $  \S = \pi\tau$, 
where $\pi \in \R^{q\times q}$ is a permutation of the identity matrix
and $\tau = [\I_l, \mathbf{0}]^{\top} \in \R^{q \times l}$ is a
truncation matrix with $\mathbf{0}$ the appropriately sized matrix
of all zeros.   It is important to note that for this particular
choice of $\Phi$, we neither explicitly construct $\S$ nor form
$\A\S$. Instead, we can randomly sample $l$ columns of the matrix $\A$
and ignore the rest. Even with other
choices of $\S$, it is never necessary to store the entire matrix $\A$
in memory as we can read in the rows sequentially to multiply by
$\S$.  Different methods for
generating the permutation matrix $\pi$ are available which are
discussed in more detail in \autoref{sec:relatedWork}.

There are alternative methods for choosing $\S$ that attempt to approximate the
column space of $\A$ by making $\S$ a random, dense matrix, such as
a subsampled randomized Fourier (or Hadamard) transform or a matrix
of i.i.d Gaussians.  The product $\A\S$ is then post-processed into an
orthogonal approximation to $V(\A)$.  See
\citet{halko2011finding} or \citet{tropp2011improved} for details.  Though these
techniques are very promising, a joint analysis of these techniques
along with the \N and \CS methods is beyond the scope of this paper
and should be addressed in future work.

For sketching matrices $  \S = \pi\tau$, we can without loss of generality suppose
there exists the following block-wise structure to the matrix $\A$
\begin{equation}
  \A = 
    \begin{bmatrix}
      \A_{11} & \A_{21}^{\top} \\
      \A_{21} & \A_{22}   \\
    \end{bmatrix}
  \label{eq:generalA}
\end{equation}
such that 
\begin{equation}
  L(\A) := \A\S = \left[ \begin{array}{c} \A_{11} \\\A_{21}
      \\ \end{array} \right] \in \R^{q \times l}.
\end{equation}
As $\A_{11}$ is symmetric, we can write its spectral decomposition as
\begin{equation}
  \A_{11} = V(\A_{11}) \Lambda(\A_{11}) V(\A_{11})^{\top}.
\end{equation}

\textbf{The \N method.}
The \N method  \citep{WilliamsSeeger2001}
uses the matrices $\A_{11}$ and $L(\A)$ to compute a low
rank approximation to $\A$ via

\begin{equation}
  \A \approx L(\A)\A_{11}^{\dagger} L(\A)^{\top}.
  \label{eq:nystromFactorization}
\end{equation}
To motivate equation \eqref{eq:nystromFactorization}, note that 
\begin{equation}
  L(\A) \A^{\dagger} L(\A)^{\top} =
  \left[
    \begin{array}{cc}
      \A_{11} & V(\A_{11}){V(\A_{11})}^{\top}\A_{21}^{\top} \\
      \A_{21}V(\A_{11}){V(\A_{11})}^{\top} & \A_{21}\A_{11}^{\dagger} \A_{21}^{\top} \\
    \end{array}
  \right].
\end{equation}
Hence, the \N method recovers the $\A_{11}$ entry exactly and a projection
of the off diagonal elements. 


To facilitate PCA on large data sets, we need to approximate the
eigenvectors and eigenvalues of $\A$ rather than attempting to approximate $\A$ itself.  
The \N method can be adapted to this purpose via the simple identity
\begin{equation}
  L(\A) \A_{11}^{\dagger} L(\A)^{\top} 
  = 
  \bigg(\rootlq{l}{q}L(\A) V(\A_{11})\Lambda(\A_{11})^{\dagger}\bigg)
  \left(\rootlq{q}{l}^2\Lambda(\A_{11})\right)
  \bigg(\rootlq{l}{q} L(\A)V(\A_{11}) \Lambda(\A_{11})^{\dagger}\bigg)^{\top},
\end{equation}
where for convenience, we define $\rootlq{c}{d} := \sqrt{c/d}$ for
$c,d\in \mathbb{N}$.
These scaling terms are somewhat crude and are intended to compensate for 
the loss of `power' incurred by subsampling.  Hence, we define the \N approximation to the
eigenvectors of $\A$ to be 
\begin{equation}
  \rootlq{l}{q}L(\A) V(\A_{11}) \Lambda(\A_{11})^{\dagger}
  =
  \rootlq{l}{q}\begin{bmatrix}
    V(\A_{11}) \\
    \Omega(\A)
  \end{bmatrix},
  \label{eq:nystromEvecs}
\end{equation}
where $\Omega(\A) := \A_{21} V(\A_{11}) \Lambda(\A_{11})^{\dagger}$,
and the \N approximation to $\Lambda(\A)$ is $\rootlq{q}{l}^2 \Lambda(\A_{11})$, 

\textbf{The Column-sampling method.}
Alternatively, we can operate on the matrix $L(\A)$ directly.  If we
decompose $L(\A)$ as  
\begin{equation}
  L(\A) = U(L) \Lambda(L) V(L)^{\top},
\end{equation}
where we suppress the dependence of $L$ on $\A$ for clarity, 
then, analogously to 
equation \eqref{eq:nystromEvecs}, the \CS approximation to the eigenvectors of
$\A$ is
\begin{equation}
  L(\A) V(L) \Lambda(L)^{\dagger} = U(L).
  \label{eq:csEvecs}
\end{equation}
Likewise, the approximate eigenvalues of $\A$ are given by $\rootlq{q}{l}\Lambda(L)$.

\subsection{Related Work}
\label{sec:relatedWork}

The \N method has recently been used to speed up kernel algorithms in
the machine learning 
community
\citep{DrineasMahoney2005,BelabbasWolfe2009,WilliamsSeeger2001,TalwalkarKumar2008}.  
These works, in contrast with this
paper, provide theoretical or empirical bounds 
on the difference between the kernel matrix and its
approximation generated either by the \N or \CS methods.  
As the matrices $\X^{\top}\X$ and $\X\X^{\top}$ are
both kernel matrices, the bounds derived in these papers apply when performing PCA.
However, when using PCA, we are rarely interested in approximating
$\X^{\top}\X$.

The sketching matrix $\S$ is at the heart of both the \N and \CS
methods, and so, theoretical analysis \citep[for
example][]{zhang2008improved,zhang2009density,liu2010learning,arcolano2010nystrom,kumar2012sampling,Gittens:2013aa}
has focused on finding good probability distributions for sampling the
columns of $\A$. Several techniques have been proposed,
some of
the most popular being: uniform sampling, deterministically
choosing the columns with largest diagonal entry
\citep[e.g.][]{BelabbasWolfe2009}, sampling with probability
proportional to $\A_{ii}$ 
\citep[e.g.][]{DrineasMahoney2005}, or sampling proportionally to the 
$\ell^2$-norms of the eigenvectors known as the leverage scores
\citep[e.g.][]{mahoney2011randomized}.  Leverage scores are much more
expensive to compute, although there are some cheap approximations
based on power methods.  There is little agreement about the benefits
of sampling schemes more complicated than the uniform method (see
\cite{Gittens:2013aa} for a recent discussion).

We do not consider the effect of choosing different $\pi$ in forming
$\S$  -- that is the effect of
different sampling schemes. As both the \N and \CS methods require the
same sketching matrix, we wish to compare these methods post
randomization.  Hence, our results are conditional on the mechanism
that selects the columns. We return to this point in \autoref{sec:discussion}.

Lastly, an alternate approach to find the SVD 
is to form an
orthonormal basis for a Krylov subspace, which, for a given matrix of
interest $\X$, an initial vector $x$, and an iteration parameter $l$,
is the column space of $\mathbf{K}_l(x) = [x, \X x, \X^2x,
\ldots, \X^{l-1}x]$.  This approach still has a computational
complexity of $O((l+s)np)$, where $s$ is an oversampling parameter
needed to enhance convergence and hence is more computationally
expensive than approaches based on sketches. Additionally, and most seriously, it requires storing
the entire matrix $\X$ in fast memory or making repeated calls to slow
memory.  This requirement rules out the
analysis of many interesting, dense data sets.

\subsection{Our Contribution}
The literature on the \N and \CS methods centers on making operator or
Frobenius norm bounds on the difference between
$\inner := n^{-1}\X^{\top}\X$ (or alternatively $\outer := \X\X^{\top}$) 
and the approximations $L(\inner) \inner_{11}^{\dagger} L(\inner)^{\top}$ (\N method)
or $U(L(\inner)) \Lambda(L(\inner)) U(L(\inner))^{\top}$ (\CS method).
While this is important in some cases,
PCA-based applications demand techniques and bounds involving the individual
matrices of interest $V$, $U$,  and $U\Lambda$.  

While upper bounds for the distances between these targets and the
related approximate quantities
can be derived from operator norm (though not Frobenius norm) bounds \citep[for
example]{karoui2008operator},
such results tend to have at least two
problems.  First, they are much looser even than the original upper
bounds, as operator norm bounds ensure uniform closeness of 
all unit norm vectors rather than just those desired.  Second, letting
$V_{nys}$ and $U_{nys}$ be the \N approximation  
to $V$ and $U$, respectively, the previous literature cannot address
important questions such as whether $\X V_{nys}$
is a better approximation to $U_d\Lambda_d$ than $U_{nys}
\Lambda_{nys}$, where $\Lambda_{nys}$ approximates $\Lambda$.

In this paper, we produce new bounds for the information loss incurred by
a data analyst who wishes to undertake a PCA-based analysis but is
forced to perform an approximation using either the \N or \CS methods.  We
additionally propose and compare different approximations to $U$ 
or $V$.
These bounds,
along with numerical experiments, provide guidance on trading the
costs and the computational benefits of these approximation
methods in common data analysis scenarios.

%


\section{Approximate PCA}
We decompose $\X$ as follows: $\x_1 = \X\S$ and
$\X_1^{\top} = \X^{\top}\S$,
$
  \X 
  = 
  [   \X_1^{\top},      \X_2^{\top}]^{\top} 
  = 
  [\x_1, \x_2]$,
where 
$\x_1 \in \R^{n \times l}, \x_2 \in \R^{n \times (p-l)}, \X_1 \in
\R^{l \times p}$, and $\X_2 \in \R^{(n-l) \times p}$.   
To approximate $V$ via the \N and \CS methods, we form
 $\inner := n^{-1}\X^{\top}\X = n^{-1}V
\Lambda^2 V^{\top}$.  Following equations
\eqref{eq:nystromEvecs} and \eqref{eq:csEvecs}, the \N approximation
to $V$ is
\begin{equation}
  V_{nys} 
   := 
 \rootlq{l}{p} L(\inner) V(\inner_{11}) \Lambda(\inner_{11})^{\dagger} 
   = 
   \rootlq{l}{p}
    \begin{bmatrix}
    V(\inner_{11}) \\
    \Omega(\inner)
  \end{bmatrix},
  \label{eq:Vnys}
\end{equation}
where $\Omega(\inner)$ is as in equation \eqref{eq:nystromEvecs}.
As $\inner = n^{-1} \X^{\top}\X$, we see that $n\inner_{11} =
\x_1^{\top}\x_1$ and $nL(\inner) = \X^{\top}\x_1$. Therefore,
$V(\inner_{11}) = V(\x_1)$ and $n\Lambda(\inner_{11}) =
\Lambda(\x_1)^2$, so that
\begin{align}
  V_{nys}
  & =  \rootlq{l}{p} L(\inner) V(\inner_{11}) \Lambda(\inner_{11})^{\dagger} \notag \\
  & =   \rootlq{l}{p}\X^{\top}\x_1 V(\x_{1}) \Lambda(\x_{1})^{2\dagger}  \notag \\
  & =   \rootlq{l}{p}\X^{\top} U(\x_1) \Lambda(\x_1)^{\dagger}    \label{eq:VnysExpansion}\\
  & = \rootlq{l}{p} 
  \begin{bmatrix} 
  V(\x_1) \\ 
  \x_2^{\top} U(\x_1)\Lambda(\x_1)^{\dagger} 
  \end{bmatrix} \notag 
\end{align}
and $\Lambda_{nys} = \rootlq{p}{l}^2\Lambda(\inner_{11}) = n^{-1}\rootlq{p}{l}^2\Lambda(\x_1)^2$.
Thus, we can find the \N approximation to $V$ by forming $\x_1$,
calculating its
SVD, and then mapping it into the correct space via the adjoint of $\X$.
Likewise, the \CS approximations to $V$ and $\Lambda$ are
\begin{equation}
  V_{cs} := U(L(\inner)) \quad \textrm{and} \quad \Lambda_{cs} := \rootlq{p}{l}\Lambda(L(\inner)),
\end{equation}
respectively.  
Here, we see that the \CS method orthogonalizes the range of $L(\inner)$, 
which is a subset of the range of $\inner$.


For approximating $U$, 
write $\outer = \X\X^{\top} = U \Lambda^2 U^{\top} \in \mathbb{R}^{n \times n}$,
then we can apply the \N and \CS methods to $\outer$ to approximate
$U$.  Specifically, writing $\outer_{11} =
U(\outer_{11})\Lambda(\outer_{11})U(\outer_{11})^{\top}$,
\begin{equation}
  U_{nys}  = 
  \rootlq{l}{n}L(\outer) U(\outer_{11})
  \Lambda(\outer_{11})^{\dagger}
  \quad \textrm{and} \quad 
  \widetilde{\Lambda}_{nys} 
  = \rootlq{n}{l}^2\Lambda(\outer) \\
  \end{equation}
  and the \CS approximations are
  \begin{equation}
  U_{cs} = U(L(\outer)),
    \quad \textrm{and} \quad 
  \widetilde{\Lambda}_{cs} = \rootlq{n}{l}\Lambda(L(\outer)).  
\end{equation}
However, as $\outer_{11} = \X_1\X_1^{\top}$ and $L(\outer) = \X
\X_1^{\top}$, we see that approximating $U$ in this manner corresponds
to subsampling {\it rows} of $\X$, which are the observations.  
This should be compared to the $V$ approximation case,
which corresponds to subsampling {\it columns} of $\X$.
As the covariates are very likely to be redundant, this
should produce a modest approximation error.  However,
as the observations are not, we are effectively
attempting to do inference with large $p$ and smaller $n$ (see \autoref{sec:empiricalResults} and
\autoref{sec:theory}).

To ameliorate this unsavory feature, there are two other possibilities for approximating $U$
that only rely on sampling covariates.  First, note that $U = \X V
\Lambda(\X)^{\dagger}$
and therefore knowing $V$ and $\Lambda(\X)$ exactly would
allow us to compute $U$ exactly.  We can define the following approximations to
$U$ based on $V_{nys}$ and $V_{cs}$
\begin{equation}
  \hat{U}_{nys} = \X V_{nys}\Lambda_{nys}^{\dagger/2}
\end{equation}
and
\begin{equation}
  \hat{U}_{cs} = \X V _{cs}\Lambda_{cs}^{\dagger/2}.
\end{equation}

A second way is to realize that $ \Ran(\x_1) \subseteq \Ran(\X)$,
where $\Ran(\A)$ is the column space (or range) of the matrix $\A$ and
the inclusion is equality if the last $p-l$ columns (after permutation) of $\X$ are
redundant. Hence, $\ran(U(\x_1))$ provides a natural  approximation to  $\ran(U)$,
suggesting the approximation  $\hat{U} = U(\x_1)$.
See \autoref{tab:methods} for a summary of these approximation methods.

Lastly, we note that approximating the principal components of $\X$ ---
that is $U\Lambda$  --- could be accomplished
via any number of approaches such as
$U_{nys}\widetilde{\Lambda}_{nys}$,  $\X V_{nys}$, 
$U_{cs}\widetilde{\Lambda}_{cs}$,  or $\X V_{cs}$.
We do not pursue investigating these approximations directly, rather we investigate 
approximating $V$ and $U$ as a proxy.

\begin{table}[!t]
  \centering
  \begin{tabular}{c | l | l }
    Quantity of interest  & Label & Approximations \\
    \hline
    && \\
    $V$ & $V_{nys}$ & $L(\inner) V(\inner_{11}) \Lambda(\inner_{11})^{\dagger}$ \\
    & $V_{cs}$   & $U(L(\inner))$ \\
    && \\
    \hline
    && \\
    & $U_{nys}$&$L(\outer) V(\outer_{11}) \Lambda(\outer_{11})^{\dagger}$ \\
    & $U_{cs}$  &$U(L(\outer))$ \\
    $U$ &  $\hat{U}_{nys}$ & $\X V_{nys}\Lambda_{nys}^{\dagger/2}$ \\
    &  $\hat{U}_{cs}$   &  $\X V_{cs}\Lambda_{cs}^{\dagger/2}$ \\
    &  $\hat{U}$            &  $ U(\x_1)$ \\
    && \\
    \hline
  \end{tabular}
  \caption{Summary of approximation methods}
  \label{tab:methods}
\end{table}

\section{Computations}
\label{sec:computing}

The \N method for approximating $V$ can be computed in a number of ways, based on whether space 
is a limiting quantity. Two such methods are shown in 
\autoref{alg:spaceNystrom} and \autoref{alg:stableNystrom}. The first uses the
matrix $\inner_{11}$ while the second does not. In each case, the
first step is to form $\x_1=\X\S$. While computing $\X\S$
involves a large matrix multiplication, such a step is unnecessary for
either method here. In both cases, we choose a random size $l$ subset
of the integers $\{1,\ldots,p\}$ and select these columns of $\X$ to
read into memory. For more general $\Phi$, we can sequentially read in
rows of $\X$ and multiply by $\S$, and hence
never read in the entire matrix $\X$.

\begin{algorithm}[t]
  \footnotesize
  \DontPrintSemicolon
  \Input{Approximation parameter $l$}
  \BlankLine
  Form $\x_1$ by randomly selecting $l$ columns of $\X$\;
  Set $\inner_{11} = n^{-1} \x_1^{\top}\x_1 $ \;
  Compute $V(\inner_{11})$ and $\Lambda(\inner_{11})$
  \BlankLine
  \Return{$L(\inner)V(\inner_{11})\Lambda(\inner_{11})^{\dagger}$}
  \caption{Space-efficient computation of $V_{nys}$}
  \label{alg:spaceNystrom}
\end{algorithm}

\begin{algorithm}[t]
  \footnotesize
  \DontPrintSemicolon
  \Input{Approximation parameter $l$}
  \BlankLine
  Form $\x_1$ by randomly selecting $l$ columns of $\X$\;
  Compute $U(\x_1)$ and $\Lambda(\x_1)$ (the left singular vectors and singular values of $\x_1$)\;
  \BlankLine
  \Return{$\X^{\top}U(\x_1)\Lambda(\x_1)^{\dagger}$}
  \caption{Stable computation of $V_{nys}$}
  \label{alg:stableNystrom}
\end{algorithm}

Note that the final steps of both approaches can be performed in a
parallelizable way and do not involve 
reading the entire matrix $\X$ into memory at the same time.
Hence, \autoref{alg:spaceNystrom} requires the storage of only the
matrix $\inner_{11}$, which has $l^2$ entries, 
 while \autoref{alg:stableNystrom} requires storing the matrix $\x_1$, which has $nl$ entires. 
 Forming $\inner_{11} = n^{-1}\x_1^{\top}\x_1$ and getting its eigenvector
decomposition has the same computational complexity as getting
$\Lambda(\x_1)$ and $V(\x_1)$ directly from $\x_1$ (O$(nl^2 +
l^3)$).  If space allows, \autoref{alg:stableNystrom}
is preferable as it is more stable than \autoref{alg:spaceNystrom}.

Alternatively, the \CS method
requires forming $L(\inner) \in \mathbb{R}^{p \times l}$, which has complexity O($lnp$),
and its left singular vectors and singular values, which has complexity
O($p^2l$).  
For space constraints, \N only requires the storage
and manipulation of $\inner_{11}$, which has $l^2$ entries, while \CS
requires using the $lp$ entries found in $L(\inner)$.  Therefore, there is
a substantial savings in both computations and storage when choosing the \N method
over the \CS method.  See \autoref{tab:methodsComplexity}
for a summary of these complexities.
Lastly, if an approximation to $U$ is desired, $\hat{U}_{nys} = \X V_{nys} \Lambda_{nys}^{\dagger/2}$ 
or  $\hat{U}_{cs} = \X V_{cs} \Lambda_{cs}^{\dagger/2}$ can be readily computed via its definition.

A significant advantage of \CS over \N is that the
columns of $V_{cs}$ are orthogonal by virtue of being the left singular vectors of $L(\inner)$.  
This imbues \CS with better
numeric properties and fewer concerns about the meaning of a
non-orthogonal approximation to $V$.  We could, in principal,
orthogonalize $V_{nys}$ to achieve a middle ground between these two
methods.  However, this orthogonalization step, performed, say, by a
QR decomposition has complexity $O(p^2l)$, which is of
the same order as the singular value decomposition of $L(\inner)$ and hence
would eliminate the computational advantage of choosing the \N method over the \CS
method to begin with.

\begin{singlespace}
\begin{table}[!h]
  \centering
  \begin{tabular}{l | l | l |}
                    & \multicolumn{2}{c|}{Complexity:} \\
Method  & Computational & Storage \\
    \hline
    \N            & O$(nl^2 + l^3)$ & O($l^2$) [O($nl$)] \\
    \CS         & O$(lnp + p^2l)$ & O($pl$)\\    
  \end{tabular}
  \caption{The complexity of the \N and \CS methods for 
  approximating $V$.  The brackets indicate
  \autoref{alg:spaceNystrom} and
  [\autoref{alg:stableNystrom}].
  In particular, the \N is only linear in $n$ and $p$ while the \CS method is quadratic.}
  \label{tab:methodsComplexity}
\end{table}
\end{singlespace}

\section{Empirical results}
\label{sec:empiricalResults}
Before turning to theoretical guarantees, we present two empirical
comparisons of the \N and \CS methods. The first is a simulation study
that explores the approximation 
accuracy.  The second is an analysis of a large corpus of
emails sent at the company Enron in the months before its collapse.

\subsection{Simulation}
\label{sec:simulation}
\textbf{Outline.}
We record four simulation conditions for comparing these approximation methods.  In all
cases, we draw $n = 5000$ observations from a multivariate normal distribution on $\mathbb{R}^p$
with zero mean and covariance matrix $\Sigma$, where $p = 3000$. 
We study four different covariance conditions labelled  $\random{0.001}$, 
$\random{0.01}$, $\random{0.1}$, and \band. Here, $\random{x}$
indicates that, with probability $x$ and for $i<j$,
$\Sigma_{ij}^{-1} = 1$ and 0 otherwise (diagonal 
elements are always equal to 1) and $\Sigma^{-1}$ is further symmetrized so that for $j > i$, 
$\Sigma_{ij}^{-1}$ is set to the same value as
$\Sigma_{ji}^{-1}$. Likewise, \band\ indicates that $\Sigma_{ij}^{-1} =
1$ if $|i-j| \leq 50$ and 0 otherwise. 
For the graph generated by these precision matrices, these simulation conditions result in approximately 
$\binom{p}{2}\cdot x$ edges for $\random{x}$ and exactly $25(2p-1 - 50)$ edges for \band.  



For each simulation condition, we consider forming both $V_d$ and $U_d$ and their respective
approximations from \autoref{tab:methods} for a variety of $d$'s.
For each $d$, we compute the $d$-dimensional
approximation method for 10 equally spaced $l$ values between $3d/2$ and $\min\{ 15d, 2p/5\}$.
We record the total computation times (in seconds) in \autoref{tab:Vnys} and \autoref{tab:Vcs} for the 
\N and \CS methods, respectively.  These computations are from {\tt R 2.15.3} on an iMac
desktop with a 2.9GHz quad-core Intel Core i5 processor and 8 gb of memory.  The PCA eigenvectors $V_d(S)$  are computed using the package
{\tt irlba} on the $\X$ matrix and we use \autoref{alg:stableNystrom} for the \N method.

\begin{table}[!h]
\centering
\resizebox{5.3in}{!}{
\begin{tabular}{r|rrrrrrrrrr|r}
$d$ & \multicolumn{10}{c}{Approximation parameter ($l$)} & \multicolumn{1}{c}{$V_d(S)$} \\
\hline
2  &  0.4 & 0.4 & 0.3 & 0.3 & 0.5 & 0.7 & 0.8 & 0.7 & 0.7 &  0.7 & 3.3 \\
3 &   0.3 & 0.2 & 0.3 & 0.3 & 0.5 & 0.7 & 0.7 & 0.8 & 0.8 &  0.8 & 4.5  \\
30 &  1.3 & 1.1 & 1.6 & 1.9 & 2.0 & 2.9 & 2.9 & 3.8 & 7.1 & 12.0	&	       29.6 \\
97 &  5.0 & 4.9 & 7.4 & 8.3 & 16.1 & 19.9 & 30.2 & 28.3 & 36.8 & 44.3	&       84.0 \\
164 & 6.9 & 7.4 & 10.4 & 11.9 & 16.7 & 19.7 & 25.8 & 30.5 & 36.7  & 43.5	&       117.0 \\
231 & 9.3 & 10.9 & 14.1 & 16.5 & 19.5 & 24.6 & 27.5 & 32.6 & 40.8 & 45.3	  &     299.0 \\
299 & 14.2 & 14.3 & 19.3 & 20.2 & 22.2 & 26.7 & 30.8 & 34.8 & 40.9 & 46.6	  &     521.5 \\
366 & 17.5 & 18.4 & 21.2 & 23.5 & 37.9 & 36.8 & 36.2 & 39.7 & 43.0 & 46.2	    &   791.6 \\
433 & 20.9 & 23.0 & 27.2 & 31.6 & 34.0 & 33.7 & 39.7 & 41.2 & 45.0 &  49.3 & 1088.0 \\
500 & 31.3 & 30.3 & 31.6 & 43.8 & 45.2 & 49.9 & 52.5 & 54.8 & 58.4  & 62.5 & 1395.3
\end{tabular}	
}						   
\caption{Computing times (in seconds) for the \N method, averaged over 4 runs: For
  each $d$, the approximation parameter, $l$, is on an equally spaced
  grid of length 10 from  $3d/2$ to $\min\{ 15d, 2p/5\}$. 
 }
\label{tab:Vnys}
\end{table}

\begin{table}[!h]
\resizebox{6in}{!}{
\begin{tabular}{r|rrrrrrrrrr|r}
$d$ & \multicolumn{10}{c}{Approximation parameter ($l$)} & \multicolumn{1}{c}{$V_d(S)$}\\
\hline
2  &   0.7  & 0.6 &  0.7  & 1.0 &  1.2  & 1.1  & 1.3 &  1.3 &  1.3 &  1.5 &     3.3  \\
3  &   0.5  & 0.5  & 0.6  & 0.7  & 0.9  & 0.8  & 2.4  & 1.8  & 1.1  & 2.5	&      4.5 \\
30  &  4.9  & 3.1 &  3.6 &  7.3  & 5.9 &  6.6 &  7.4 & 10.2 & 14.6  & 12.3	 &    29.6 \\
97  & 29.5  & 12.2 & 20.8 & 27.9 & 25.4 & 36.4 & 39.7 & 38.1 & 42.3 & 44.6	  &   84.0 \\
164 & 19.4 & 33.0 & 37.6 & 44.0 & 57.3 & 52.6 & 58.3 & 65.9 & 73.6 & 80.6	   & 117.0 \\
231 & 42.3 & 57.0 & 71.8 & 80.2 & 91.4 & 93.9 & 102.3 & 108.2 & 126.2 & 131.4 &	    299.0 \\
299 & 77.0 & 142.3 & 119.9 & 135.3 & 139.1 & 146.0 & 164.0 & 190.2 & 186.0 & 198.2	 &   521.5 \\
366 & 143.2 & 155.5 & 168.0 & 212.7 & 227.4 & 236.1 & 228.9 & 231.0 & 290.2 & 323.8	  &  791.6 \\
433 & 220.8 & 244.1 & 283.7 & 303.6 & 305.4 & 321.3 & 355.5 & 309.1 & 320.9 & 346.8 & 1088.0 \\
500 & 312.9 & 351.7 & 392.5 & 417.4 & 426.4 & 427.9 & 452.9 & 418.7 & 456.4 & 432.2 & 1395.3
\end{tabular}
}
\caption{Computing times (in seconds) for the \CS method, averaged over 4 runs: For
  each $d$, the approximation parameter, $l$, is on an equally spaced
  grid of length 10 from  $3d/2$ to $\min\{ 15d, 2p/5\}$. 
}
\label{tab:Vcs}
\end{table}

\textbf{Results.}
In each of the figures below, we display the accuracy of each
approximation method relative to the accuracy of the \CS
method. Specifically, we report the Frobenius norm error of each 
method relative to the \CS method.  For example, for recovering $V_d$
with the \N method, we plot
\begin{equation}
\frac{||V_{nys,d}({V_{nys,d}}^{\top}V_{nys,d})^{-1} {V_{nys,d}}^{\top} - V_d{V_d}^{\top}||_F}{||V_{cs,d}{V_{cs,d}}^{\top} - V_d{V_d}^{\top}||_F}.
\label{eq:figureMerit}
\end{equation}
Therefore, larger values indicate inferiority to the \CS method and smaller values indicate
superiority. We choose the Frobenius norm distance to the PCA-based projection 
as we are interested in the accuracy loss of using an approximation method instead
of the PCA basis.  

\begin{figure}[h!]
  \centering
  \begin{tabular}{cc}
 \includegraphics[width=2.2in,trim=20 35 20 50,clip]
 {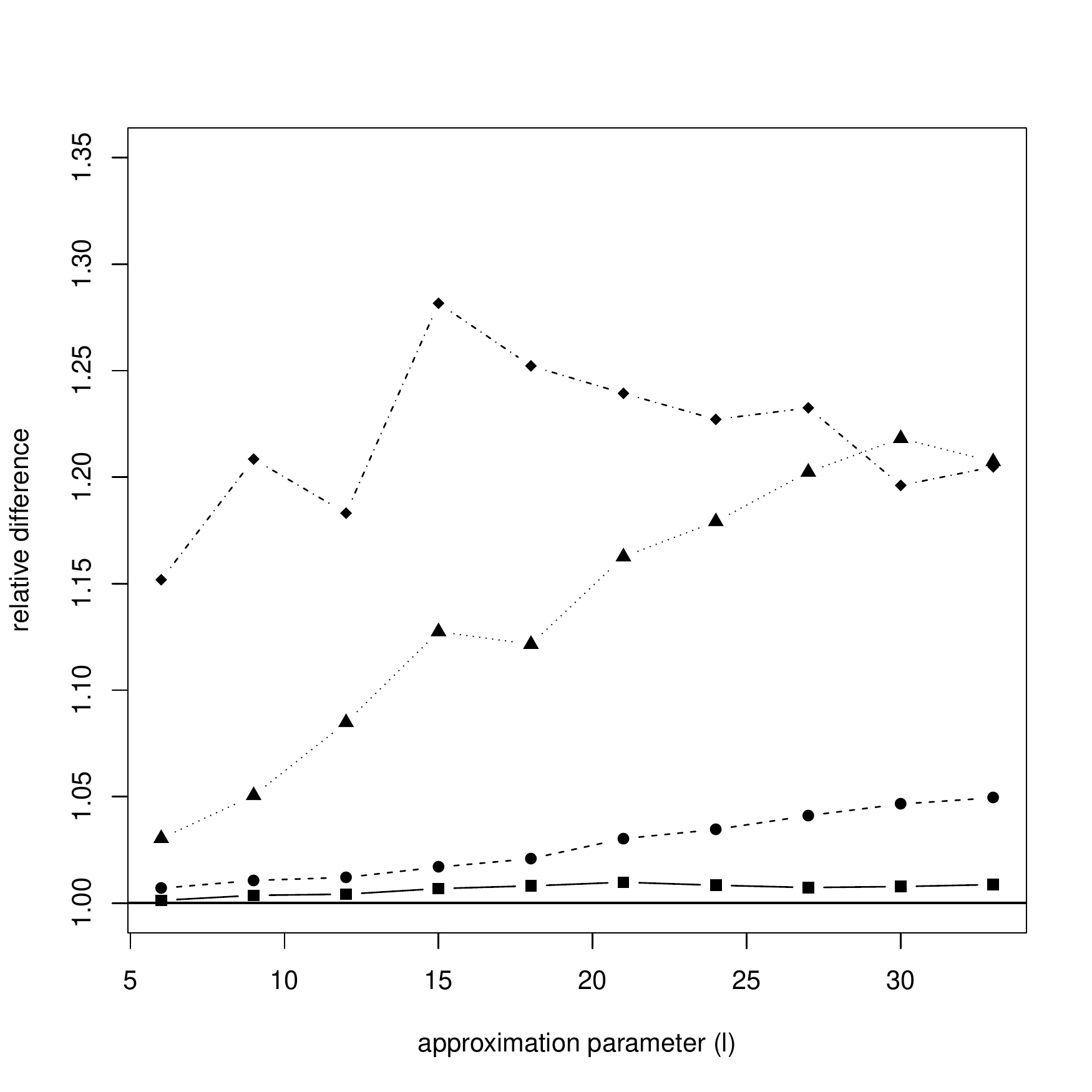}    &
 \includegraphics[width=2.2in,trim=20 35 20 50,clip]
 {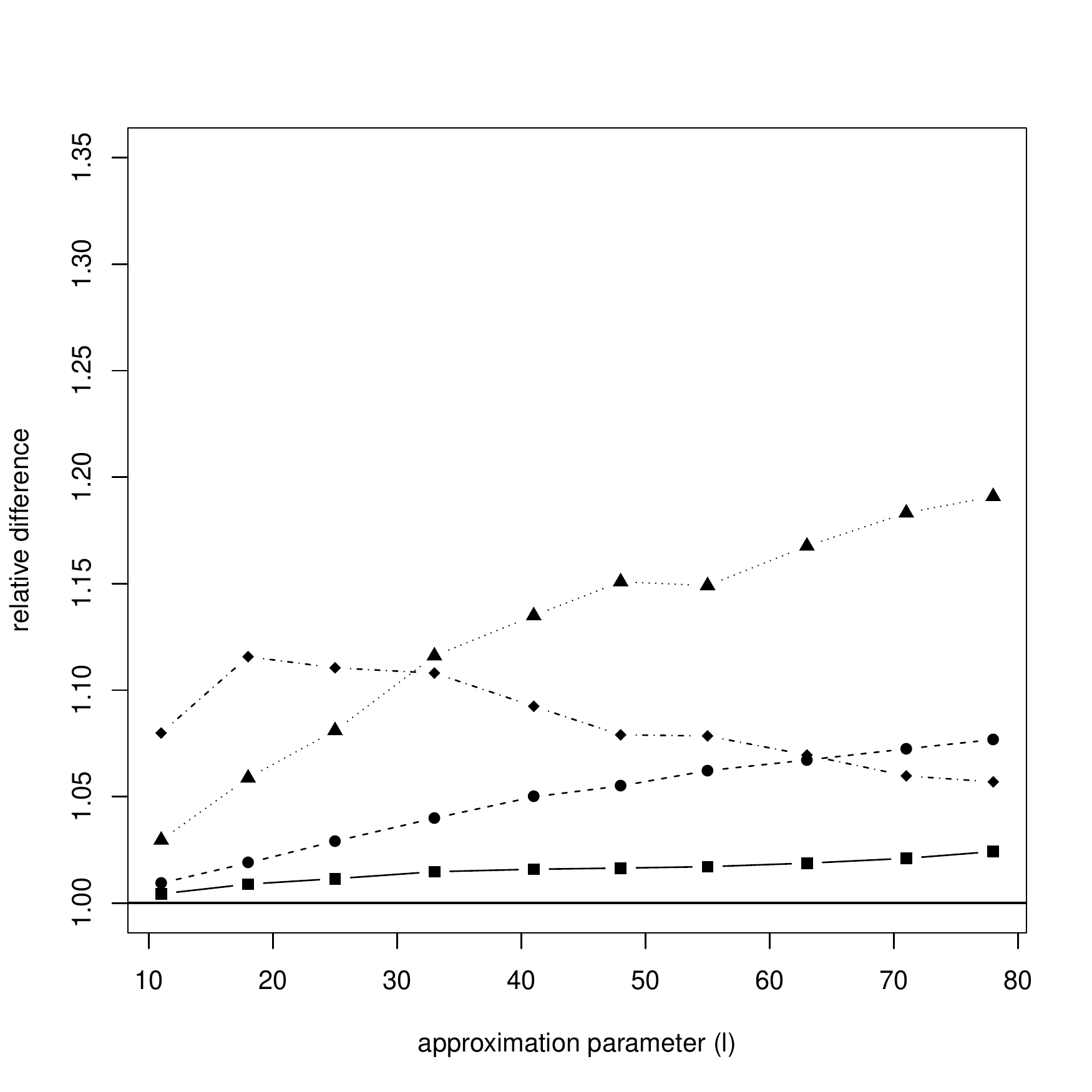}  \\
  ($d = 2$) &   ($d = 5$)    \\
 \includegraphics[width=2.2in,trim=20 35 20 50,clip]
 {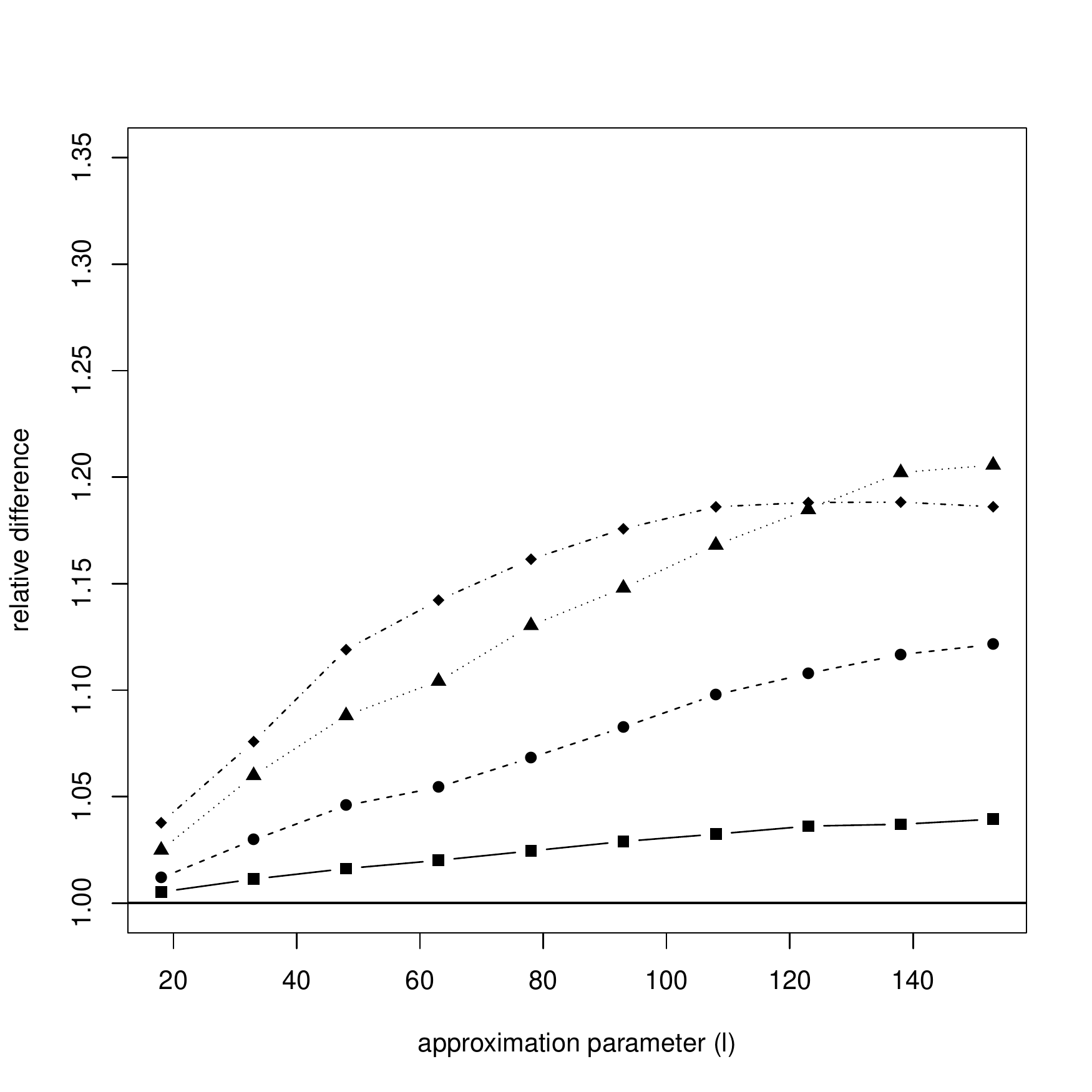}    &
 \includegraphics[width=2.2in,trim=20 35 20 50,clip]
 {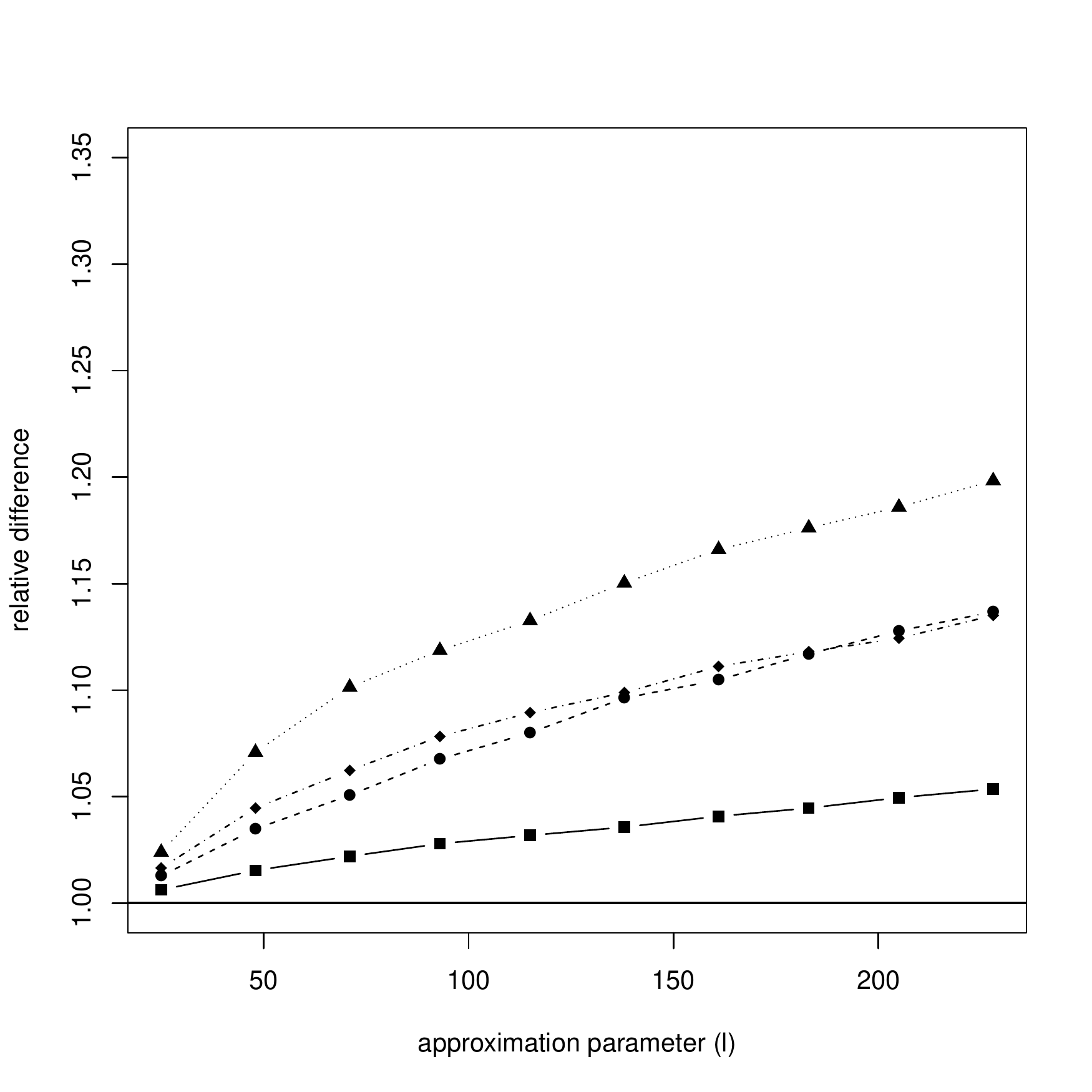}  \\
  ($d = 10$) &   ($d = 15$) \\
 \end{tabular}
\caption{This figure shows the norm difference of $V_{nys}$ to $V$ relative to the norm difference
of $V_{cs}$ to $V$ (see equation \eqref{eq:figureMerit}) 
where the $x$-axis is the approximation parameter $l$
 whose values range from $\lfloor 3d/2 \rfloor$ to $15d$. 
The four simulation conditions are $\random{0.001}$ (solid, square), $\random{0.01}$ (dashed, circle), $\random{0.1}$ (dotted, triangle), and \band (dot-dash, diamond). }
  \label{fig:simulationV}
\end{figure}

In \autoref{fig:simulationV} we plot a comparison of the \N and \CS methods when used to approximate
$V$. We see that in all cases, the \CS performs better than the \N method.
For small values of $d$, this difference is negligible for $\random{0.001}$, which is the sparsest case,
and increasingly more pronounced for the more dense cases.  The banded case demonstrates the largest
difference between the \N and \CS methods, displaying the parabolic relationship between $l$
and approximation error predicted by the theory in \autoref{sec:theory}.  
For larger values of $d$, the banded case displays a phase shift as the benefit of using
the \CS method over the \N method erodes.  This curious property is in need of further investigation.

\begin{figure}[h!]
  \centering
 \includegraphics[width=2.2in,trim=20 35 20 50,clip]
 {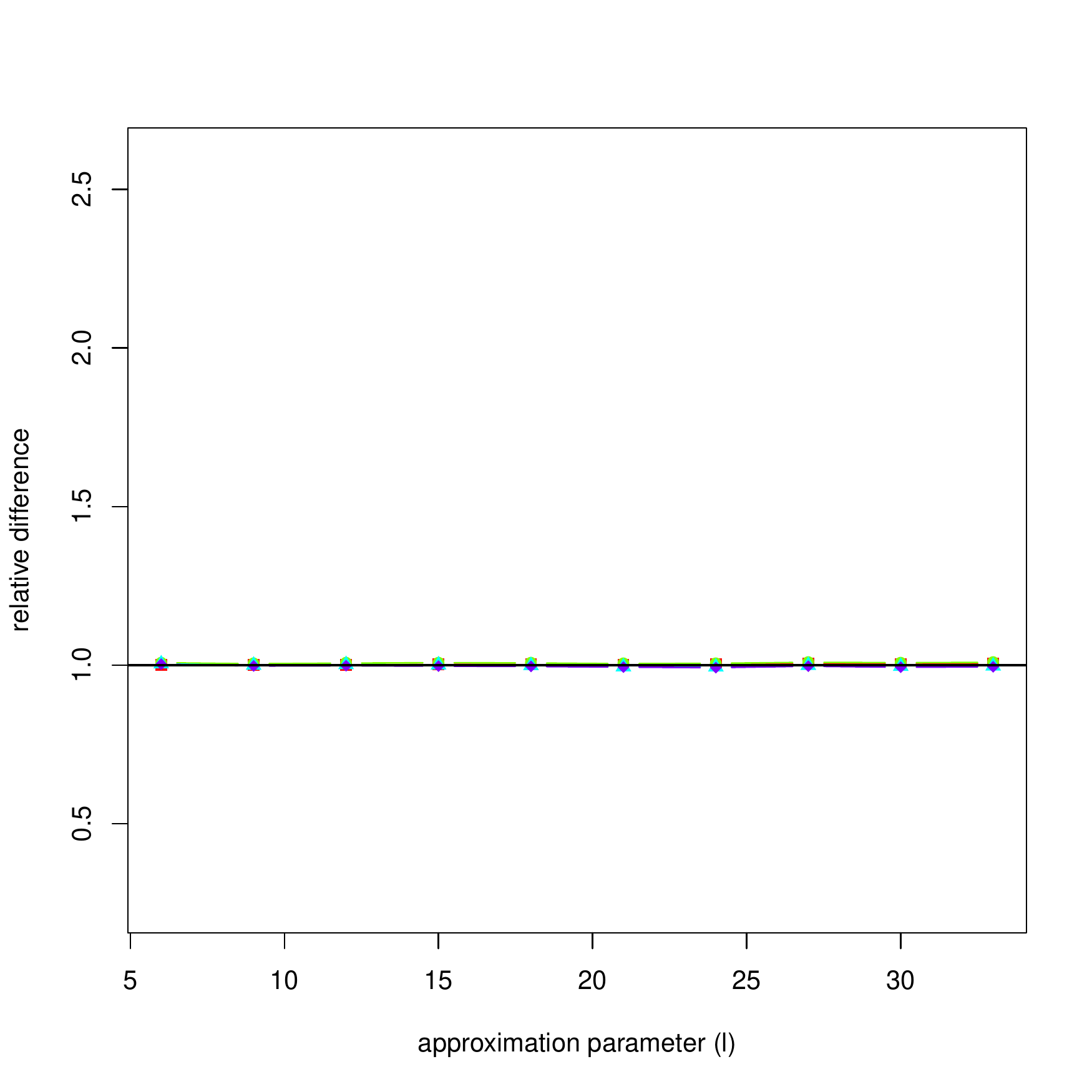}  
 \includegraphics[width=2.2in,trim=20 35 20 50,clip]
 {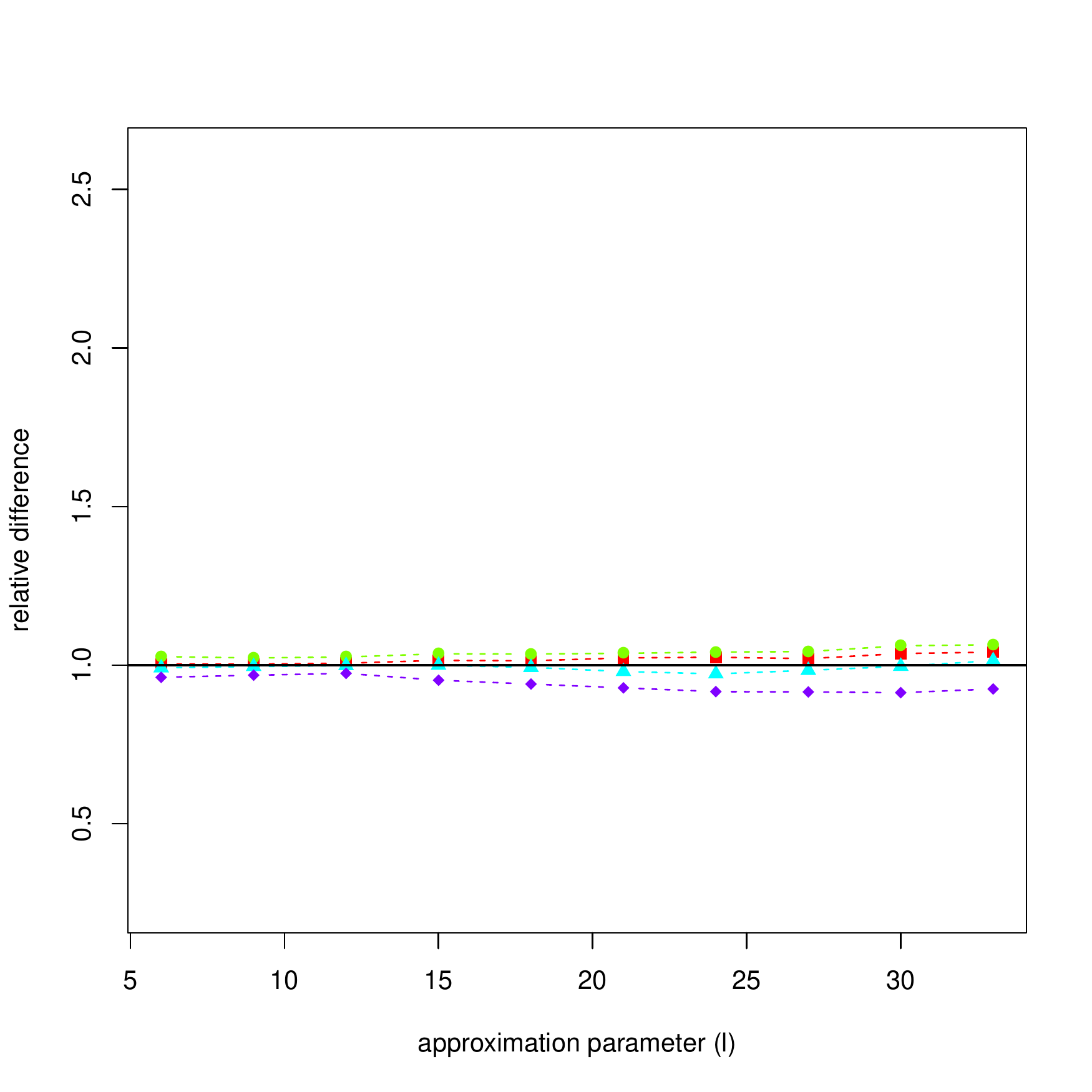}  \\
 \includegraphics[width=2.2in,trim=20 35 20 50,clip]
 {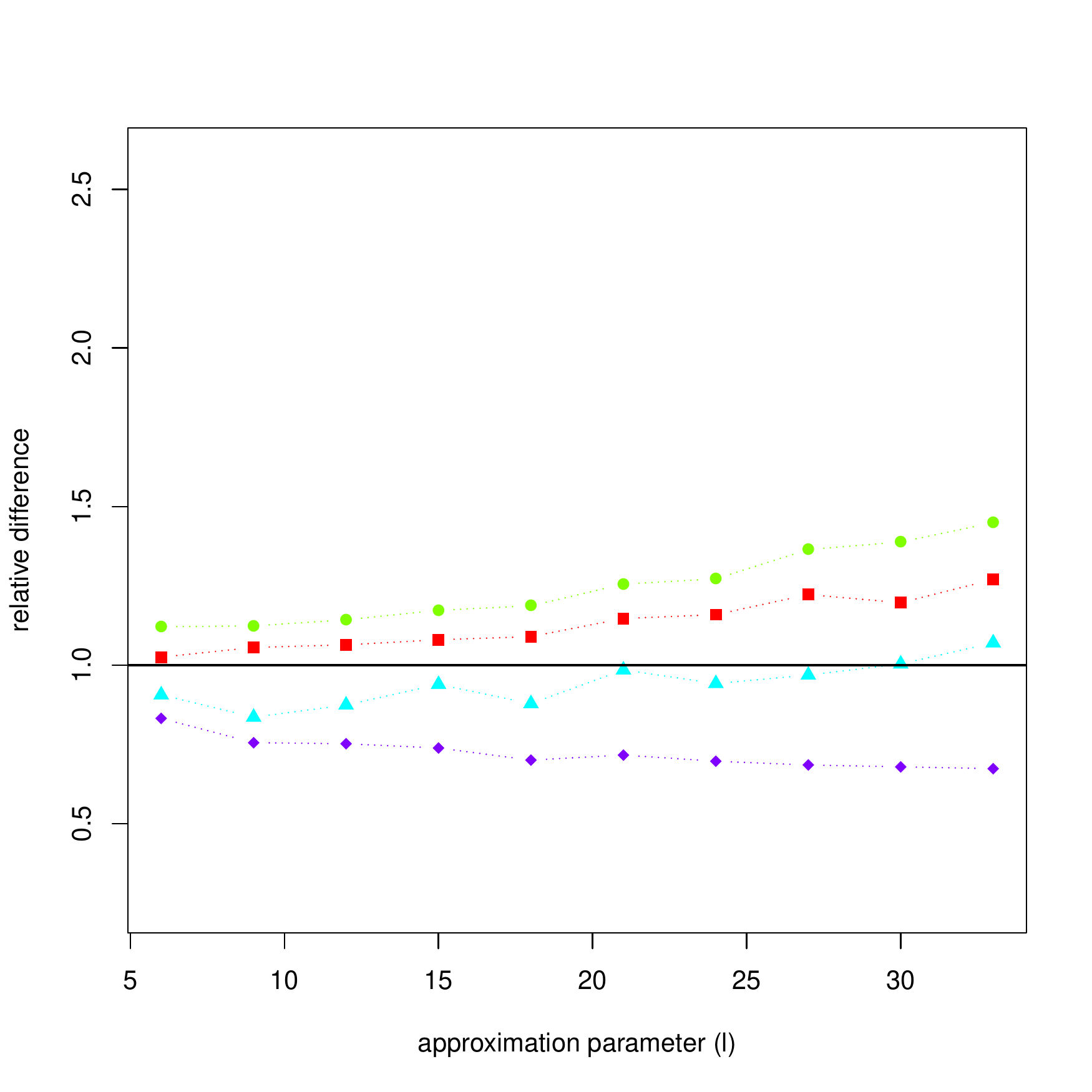}  
 \includegraphics[width=2.2in,trim=20 35 20 50,clip]
 {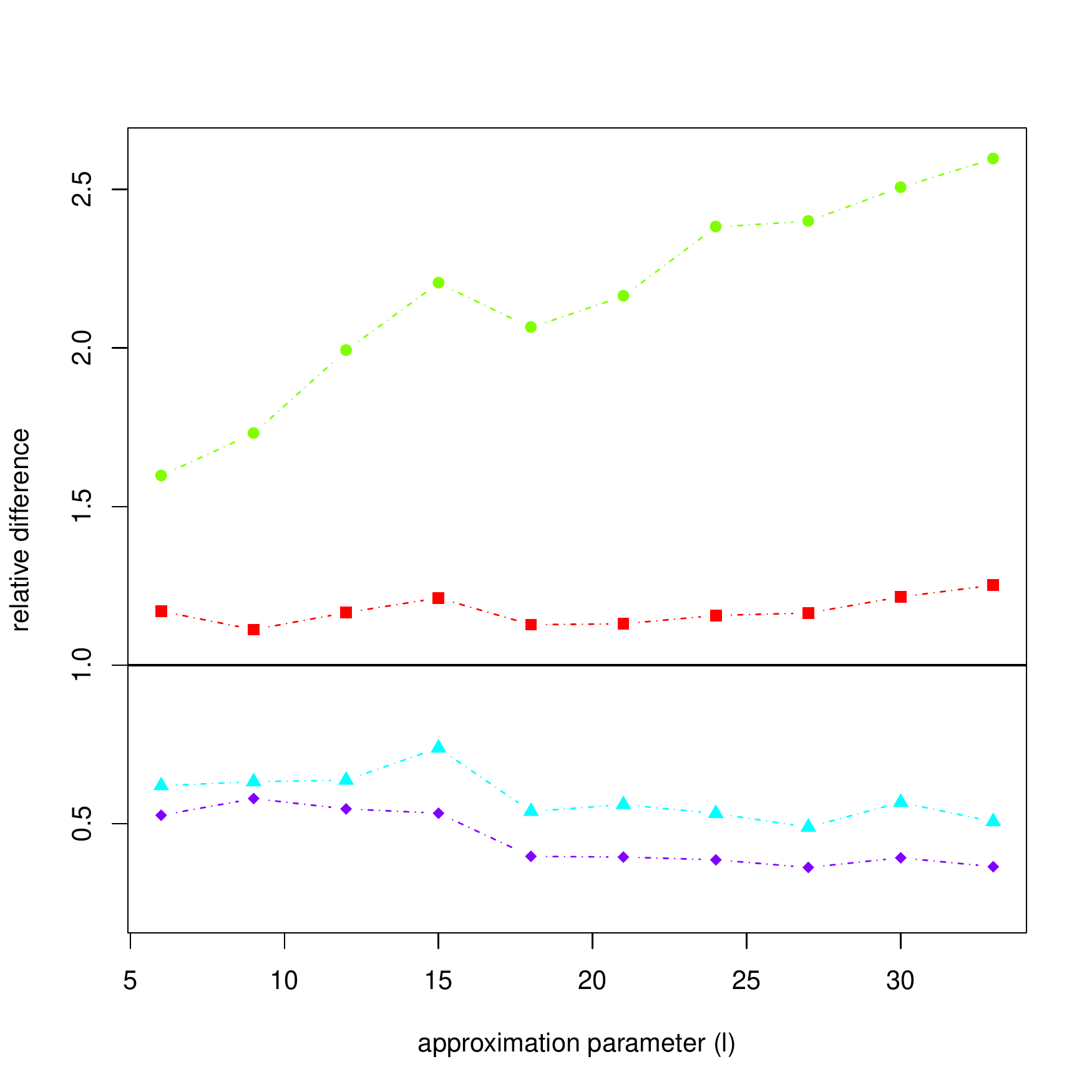}  
\caption{$d=2$. This figure shows the norm difference between [$U_{nys}$ (red square), $\hat{U}$ (green circle),
$\hat{U}_{nys}$ (blue triangle), $\hat{U}_{cs}$ (purple diamond)] and $U$ relative to the norm difference
between $U_{cs}$ and $U$ (see equation \eqref{eq:figureMerit} for an
analogous example), where 
the $x$-axis is the approximation parameter $l$. 
The four simulation conditions are $\random{0.001}$, 
  $\random{0.01}$, $\random{0.1}$, and \band\ (from top left to  bottom right).}
  \label{fig:simulationUd2}
\end{figure}

\begin{figure}[h!]
  \centering
 \includegraphics[width=2.2in,trim=20 35 20 50,clip]
 {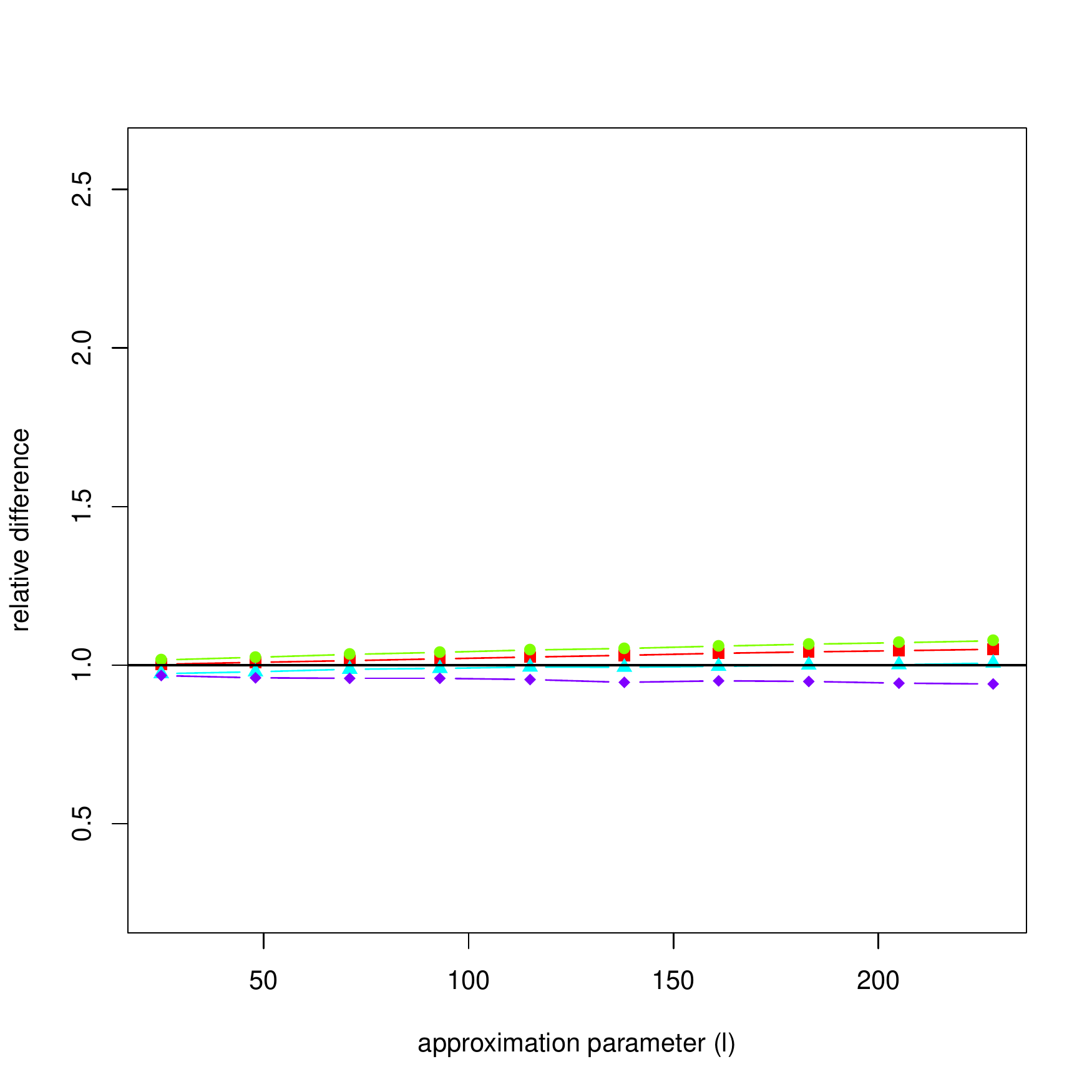}  
 \includegraphics[width=2.2in,trim=20 35 20 50,clip]
 {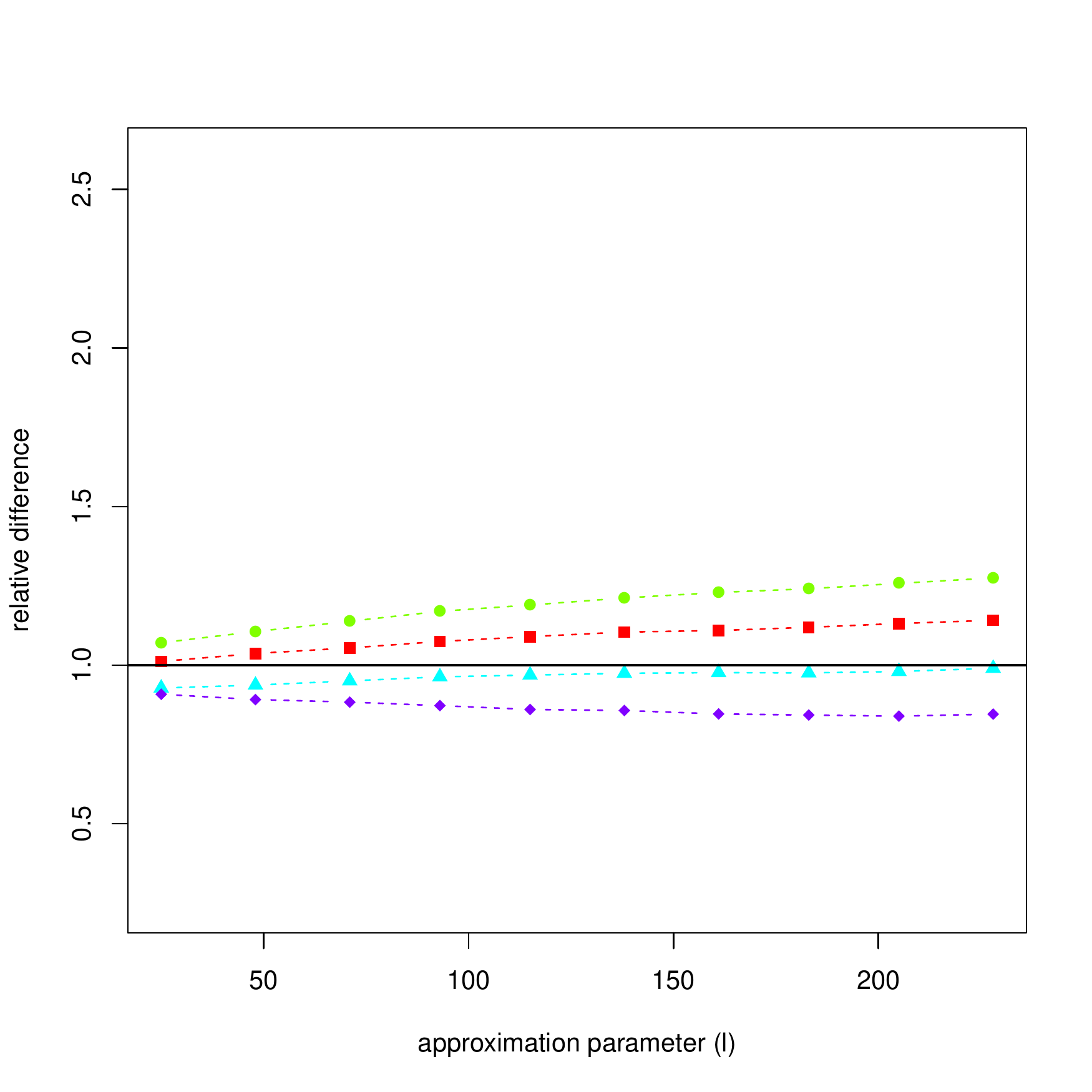}  
 \includegraphics[width=2.2in,trim=20 35 20 50,clip]
 {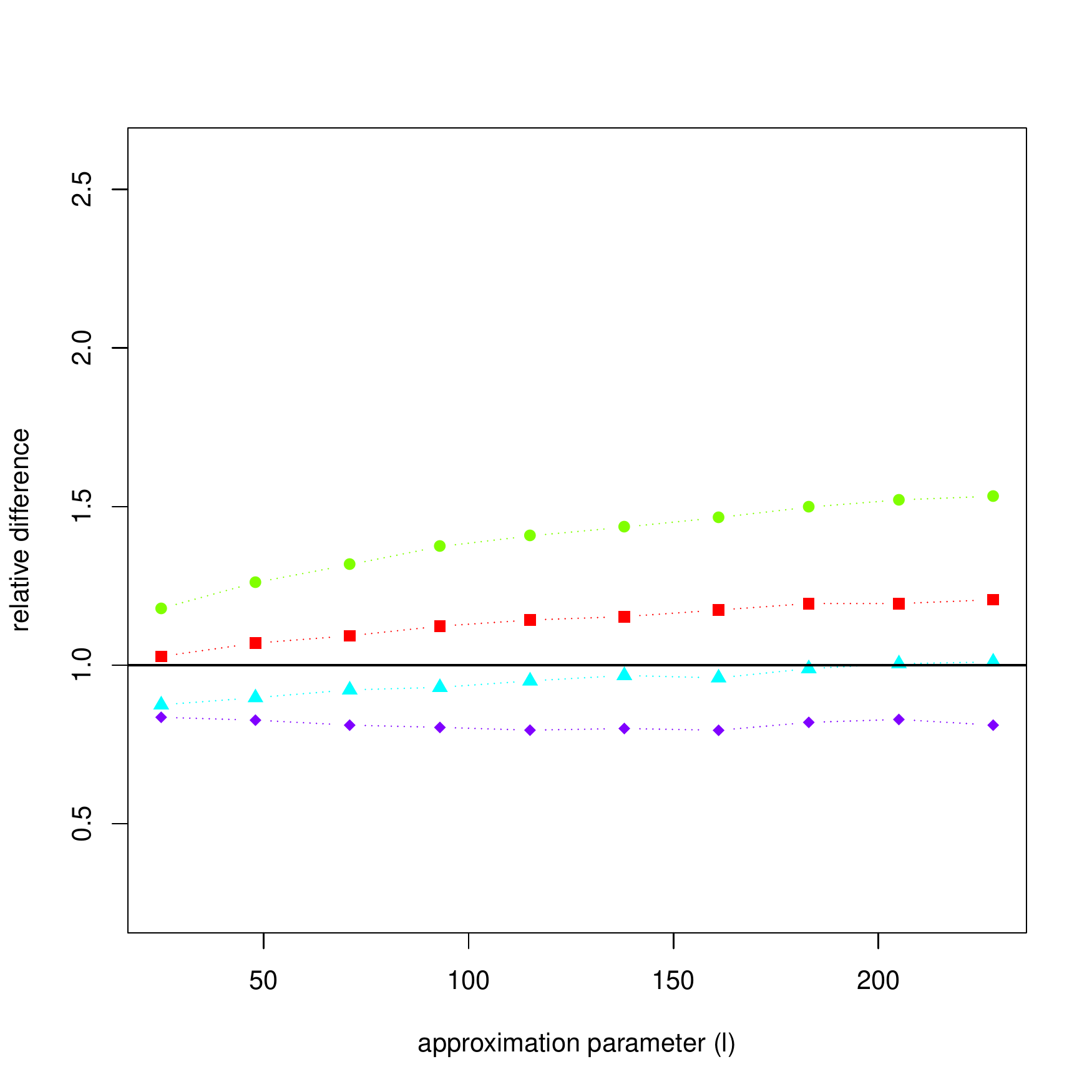}  
 \includegraphics[width=2.2in,trim=20 35 20 50,clip]
 {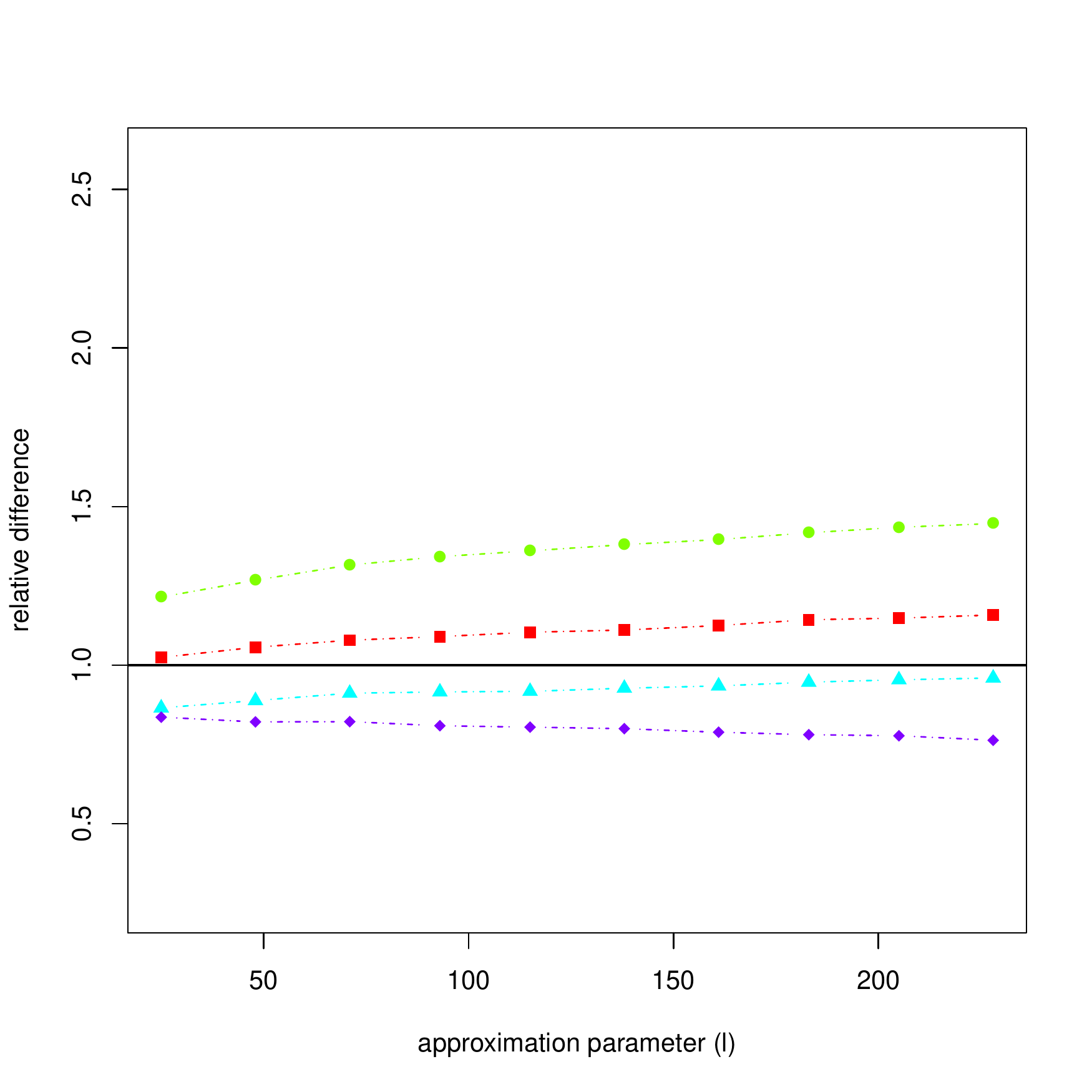}  
\caption{$d=15$. This figure shows the norm difference between [$U_{nys}$ (red square), $\hat{U}$ (green circle),
$\hat{U}_{nys}$ (blue triangle), $\hat{U}_{cs}$ (purple diamond)] and $U$ relative to the norm difference
between $U_{cs}$ and $U$ (see equation \eqref{eq:figureMerit} for an analogous example),
where the $x$-axis is the approximation parameter $l$.
The four simulation conditions are $\random{0.001}$, 
  $\random{0.01}$, $\random{0.1}$, and \band\ (from top left to  bottom right).}
  \label{fig:simulationUd3}
\end{figure}

In \autoref{fig:simulationUd2}--\autoref{fig:simulationUd3} we plot a comparison of the $U$ approximation
methods described in \autoref{tab:methods}.  For small $d$, there is almost 
no difference between the 5 methods in the $\random{0.001}$ and  $\random{0.01}$ cases 
but a profound difference for the $\random{0.1}$ and \band\ cases. As predicted, the ``plug-in''
estimators $\hat{U}_{nys}$  and $\hat{U}_{cs}$ perform better than either $U_{nys}$ or $U_{cs}$.
Somewhat surprisingly, the naive approximation $\hat{U}$ that directly approximates the column space of 
$\X$ via subsampling performs markedly worse than any of the other approximation methods.  This
observation has potentially far reaching implications  as this is a commonly used ``default'' method for
approximating the spectrum of large matrices (e.g. \citet{Bair2006prediction}).

\begin{figure}[h!]
  \centering
 \includegraphics[width=2.2in,trim=20 35 20 50,clip]
 {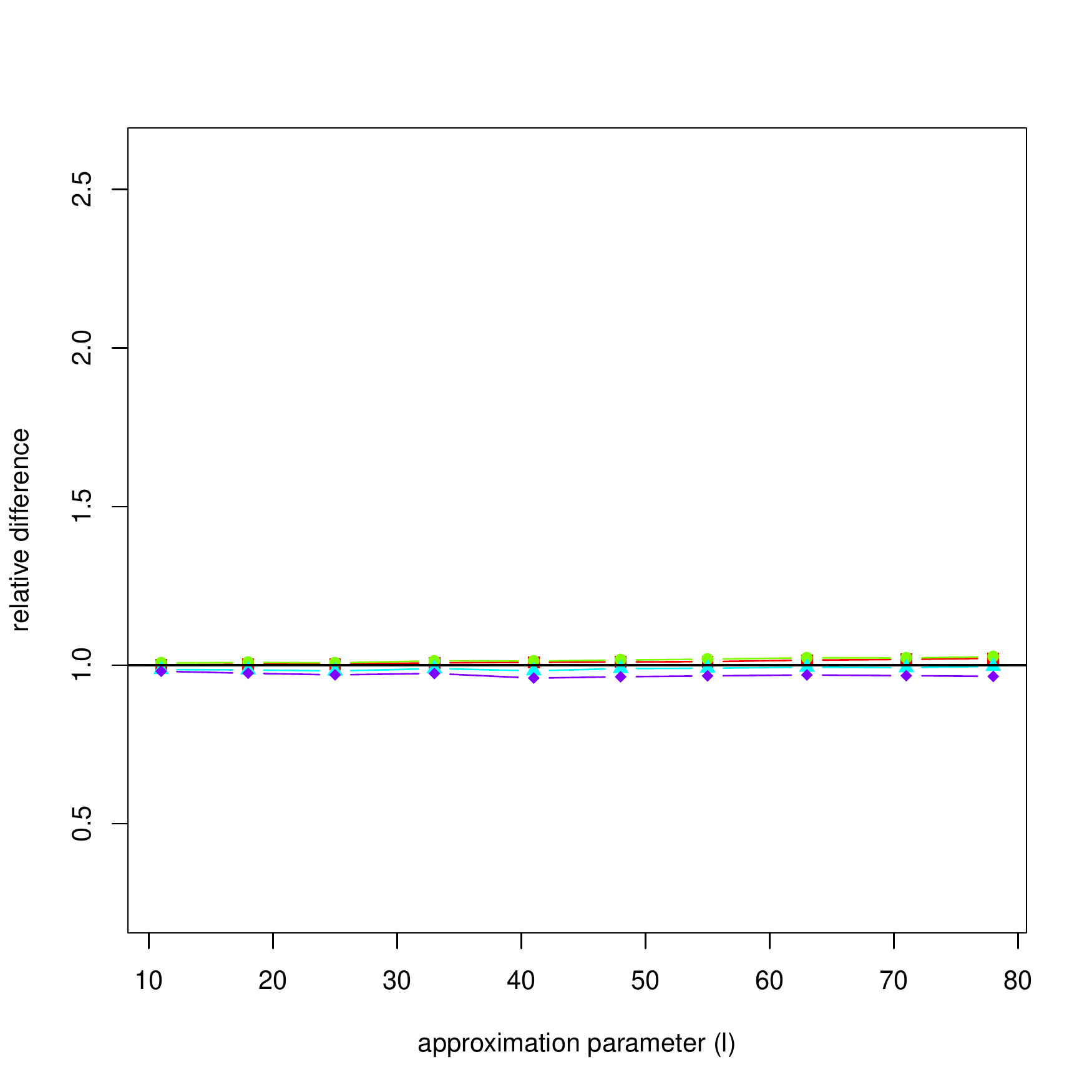}  
 \includegraphics[width=2.2in,trim=20 35 20 50,clip]
 {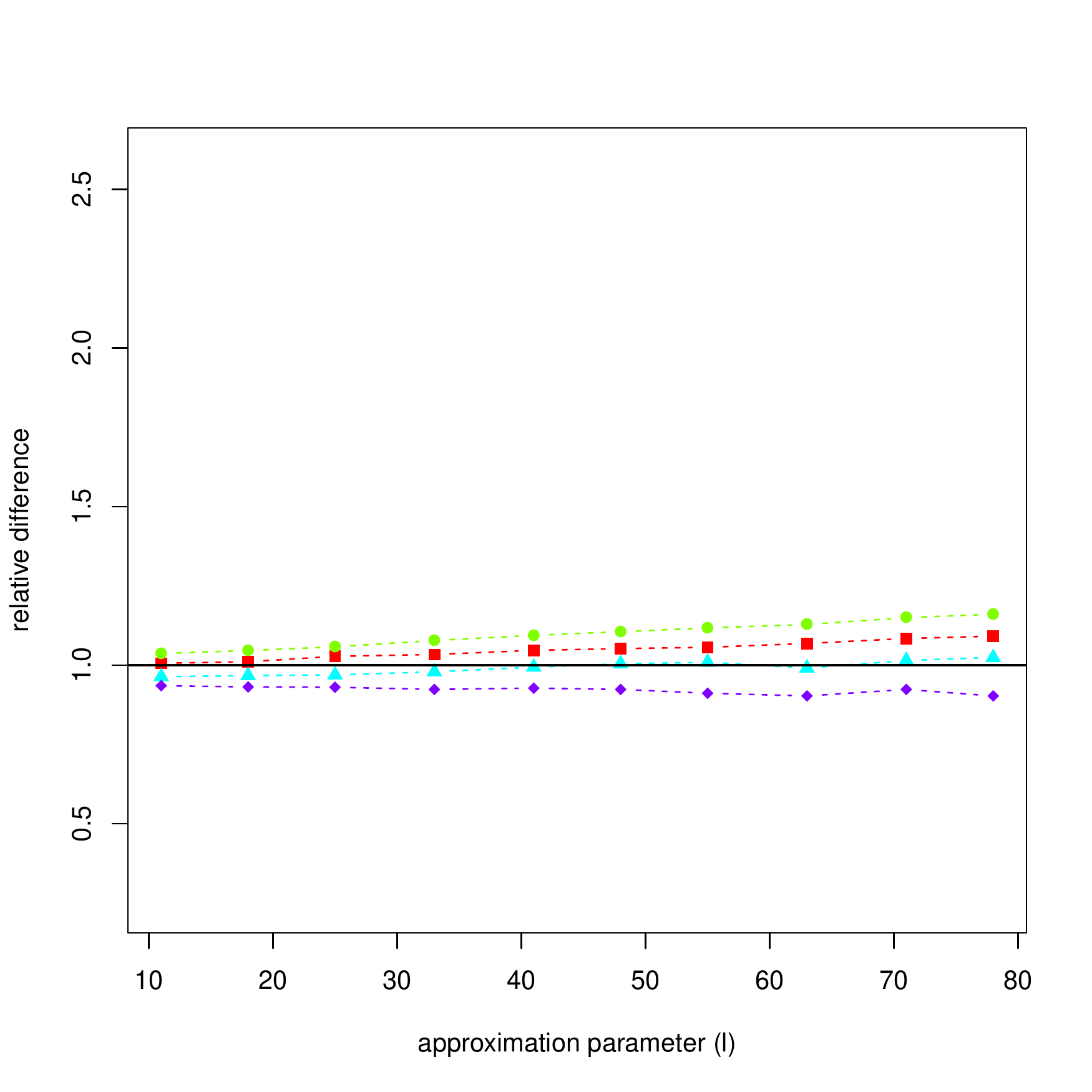}  \\
 \includegraphics[width=2.2in,trim=20 35 20 50,clip]
 {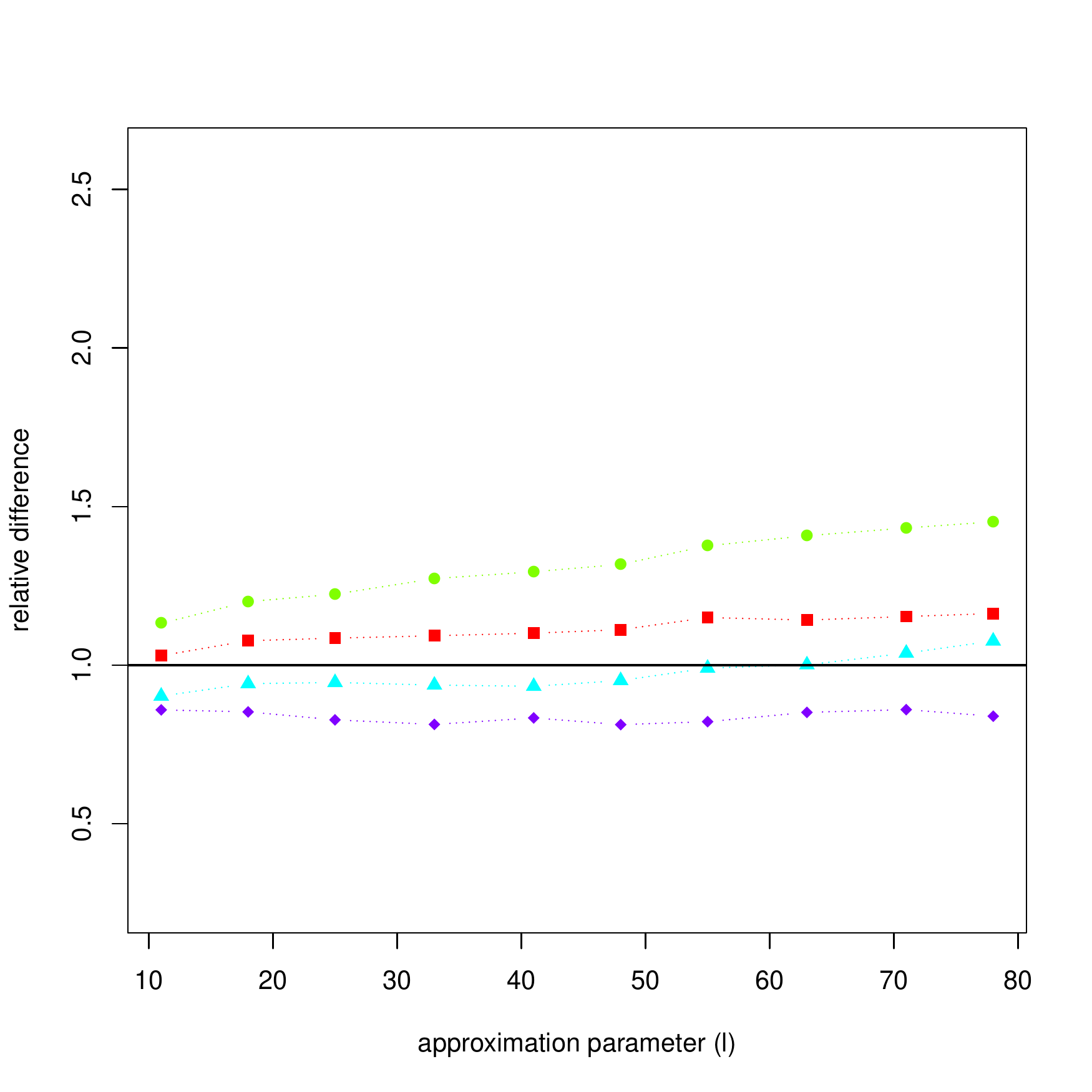}  
 \includegraphics[width=2.2in,trim=20 35 20 50,clip]
 {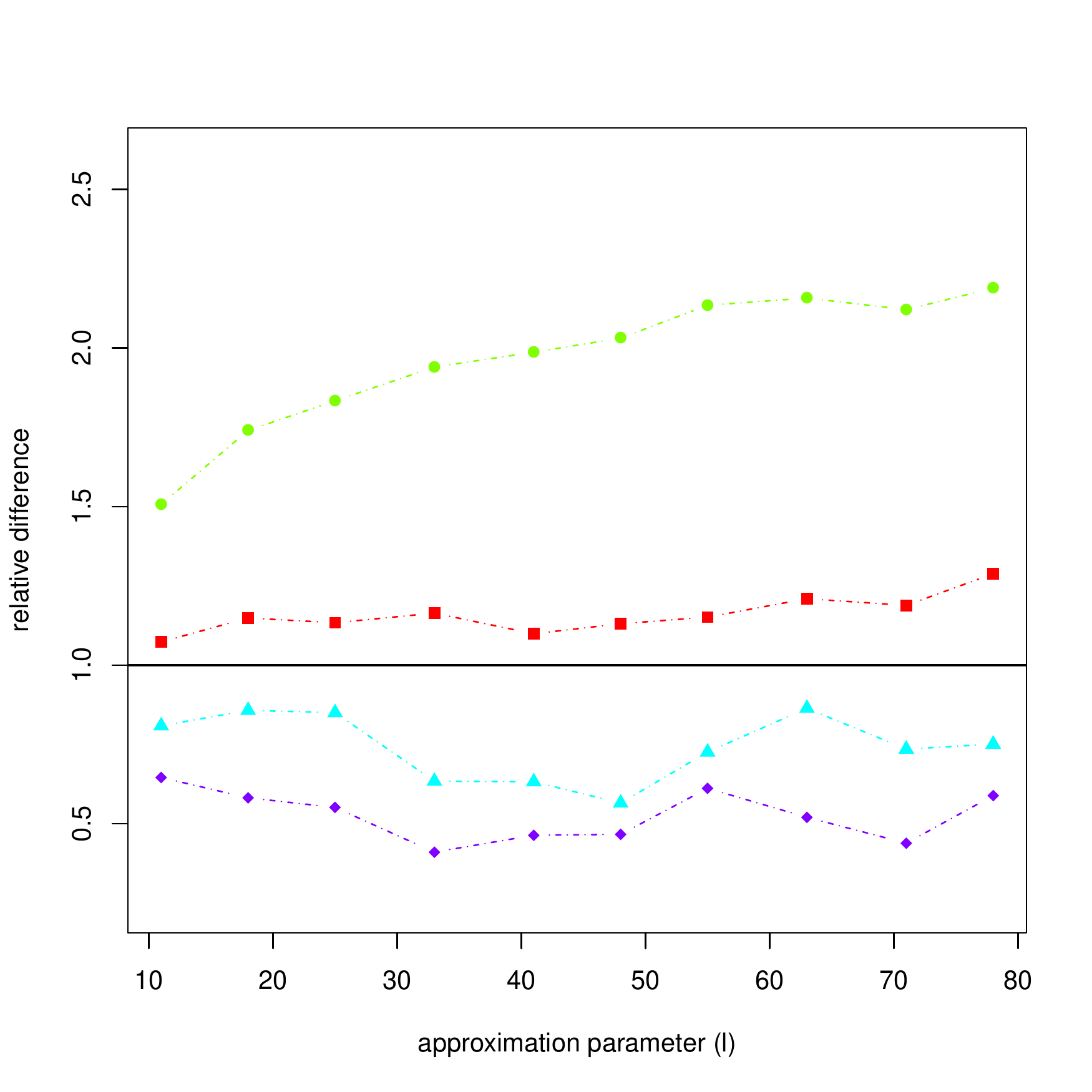}  
\caption{$d=5$. This figure shows the norm difference between [$U_{nys}$ (red square), $\hat{U}$ (green circle),
$\hat{U}_{nys}$ (blue triangle), $\hat{U}_{cs}$ (purple diamond)] and $U$ relative to the norm difference
between $U_{cs}$ and $U$ (see equation \eqref{eq:figureMerit} for an
analogous example), where
the $x$-axis is the approximation parameter $l$.
The four simulation conditions are $\random{0.001}$, 
  $\random{0.01}$, $\random{0.1}$, and \band\ (from top left to  bottom right).}
  \label{fig:simulationU}
\end{figure}

\begin{figure}[h!]
  \centering
 \includegraphics[width=2.2in,trim=20 35 20 50,clip]
 {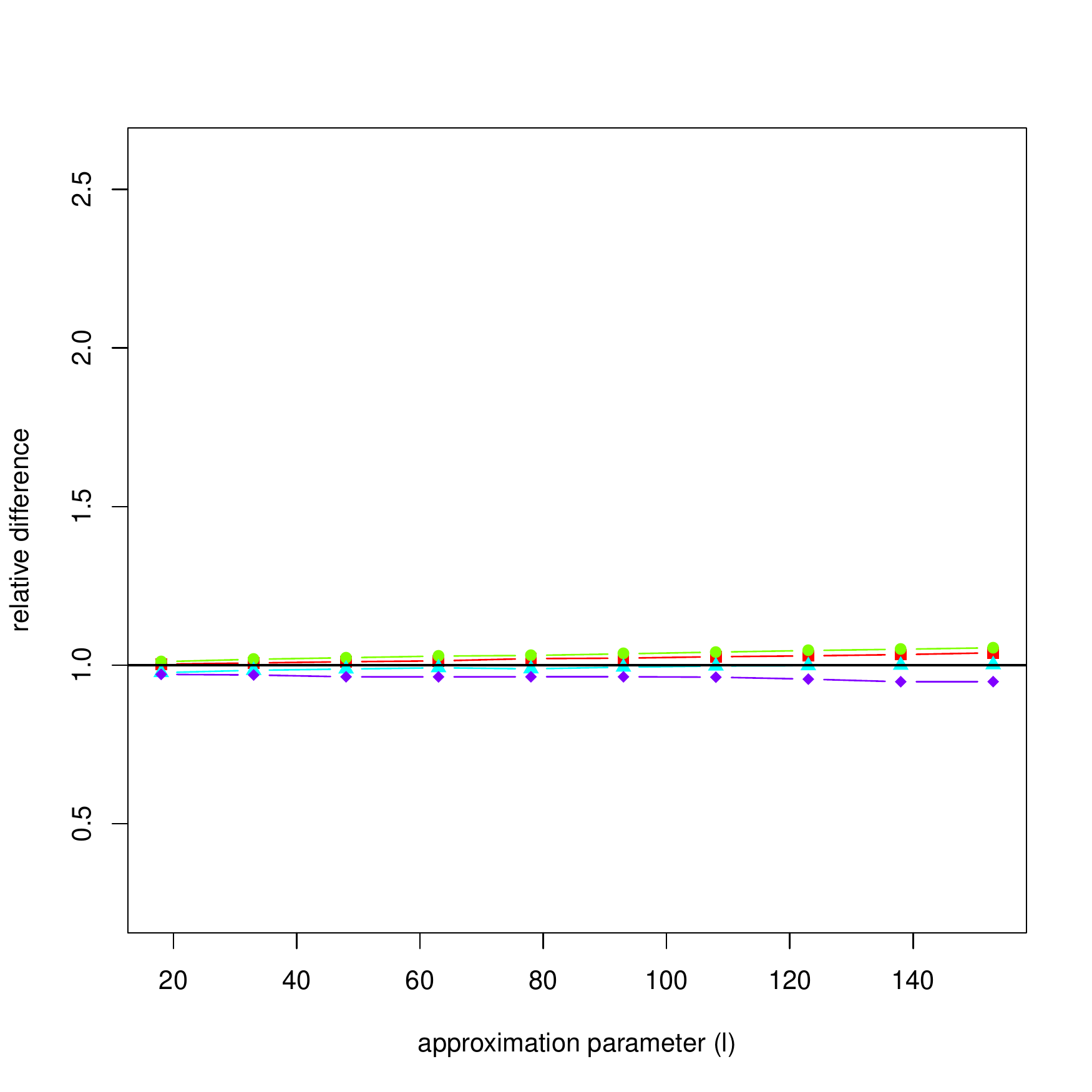}  
 \includegraphics[width=2.2in,trim=20 35 20 50,clip]
 {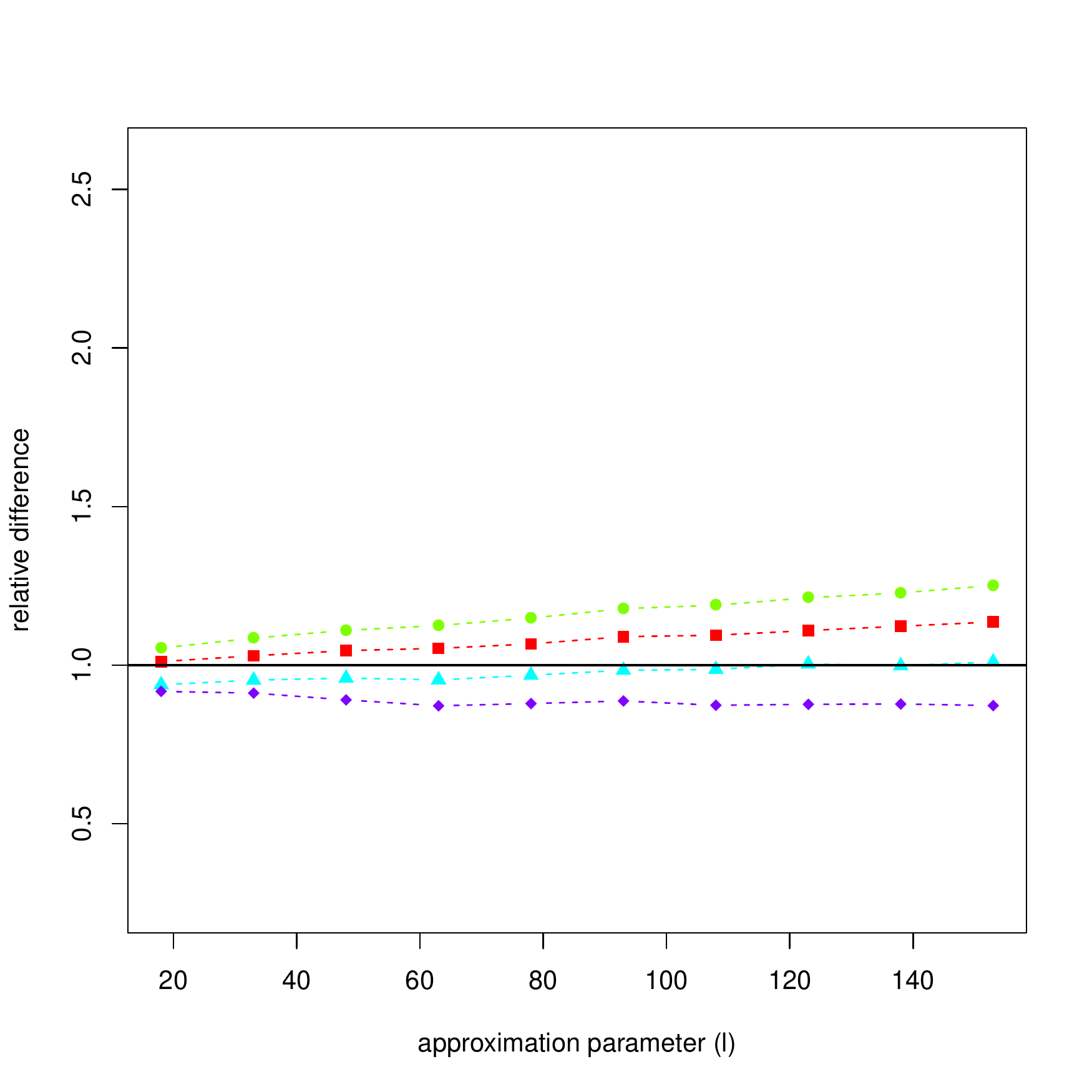}  \\
 \includegraphics[width=2.2in,trim=20 35 20 50,clip]
 {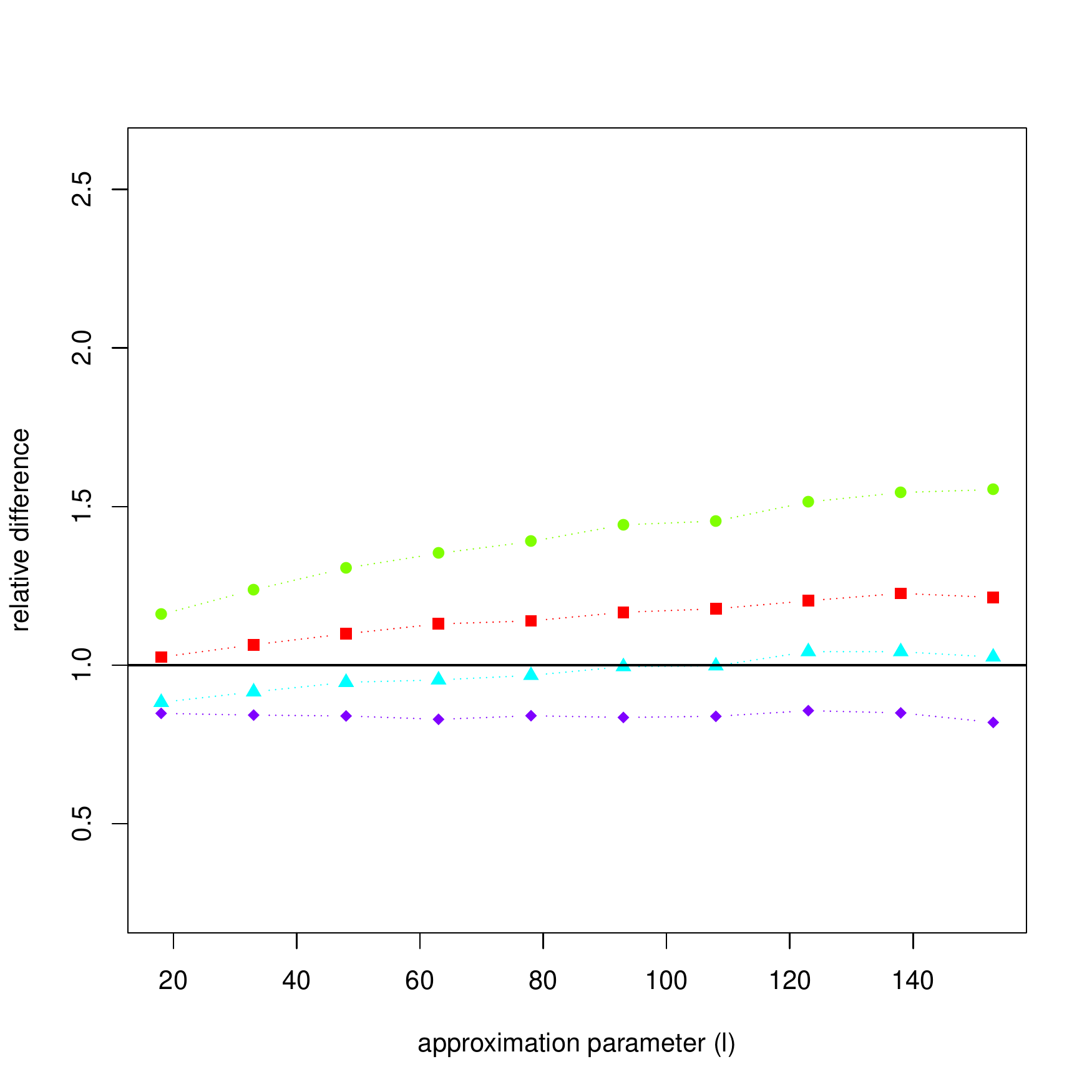}  
 \includegraphics[width=2.2in,trim=20 35 20 50,clip]
 {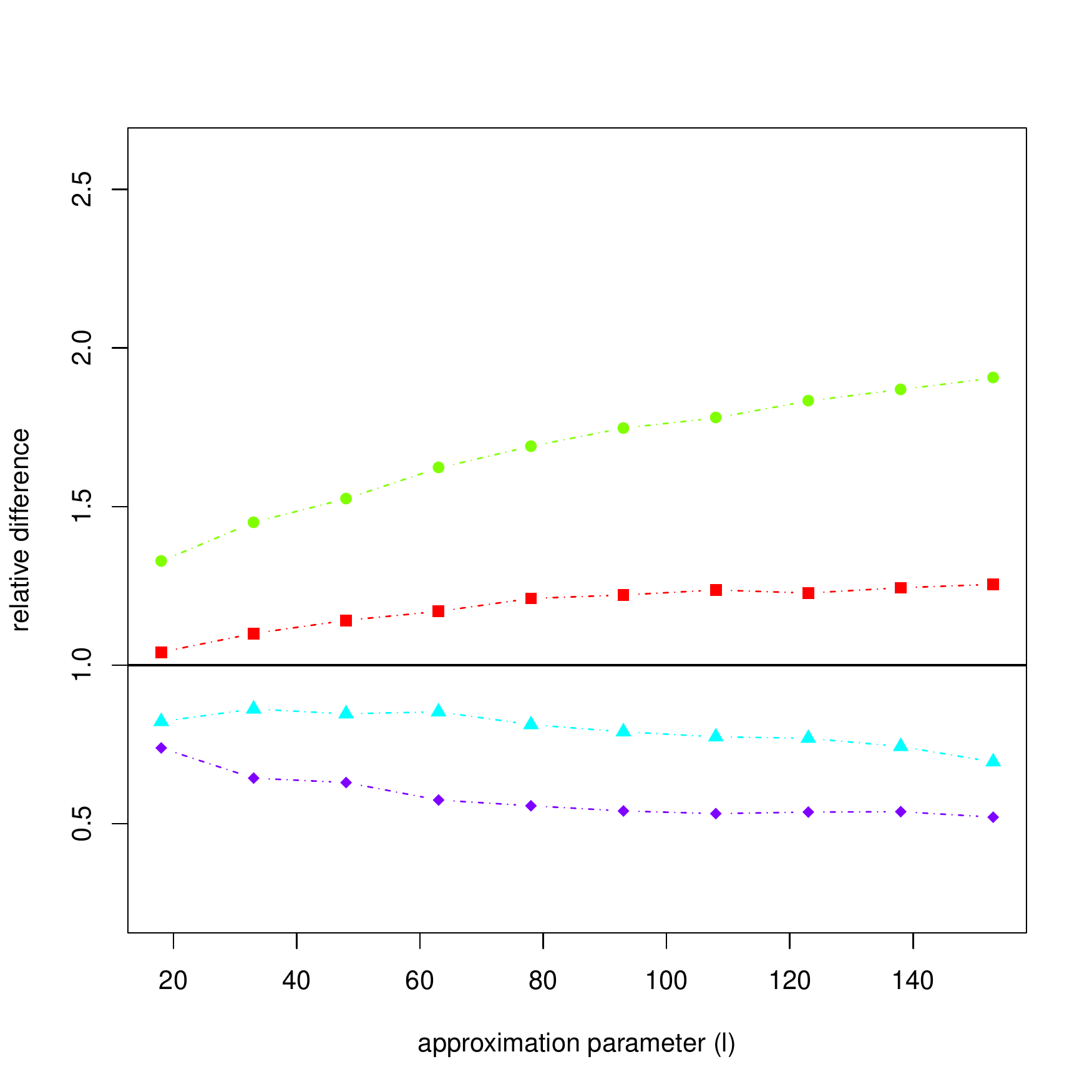}  
\caption{$d=10$. This figure shows the norm difference between [$U_{nys}$ (red square), $\hat{U}$ (green circle),
$\hat{U}_{nys}$ (blue triangle), $\hat{U}_{cs}$ (purple diamond)] and $U$ relative to the norm difference
between $U_{cs}$ and $U$ (see equation \eqref{eq:figureMerit} for an analogous example),
where the $x$-axis is the approximation parameter $l$.
The four simulation conditions are $\random{0.001}$, 
  $\random{0.01}$, $\random{0.1}$, and \band\ (from top left to  bottom right).}
  \label{fig:simulationU}
\end{figure}

\subsection{Enron data}
\label{sec:enron}

A well known text processing data set is a compilation of emails
sent between 158 employees of the energy trading company Enron 
in the months precipitating its collapse in 2001 (See
  \url{https://www.cs.cmu.edu/~enron/} for details).
After applying standard text preprocessing techniques, 
the resulting data set contains $n=39,\!861$ documents and $p=28,\!102
$ total words recorded as counts in a document-term matrix, $\tilde{\X}$.
Frequently, researchers wishing to analyze this matrix would perform
latent semantic indexing (LSI), which amounts to computing
the leading right singular vectors of the uncentered matrix $\tilde{\X}$.
Since $\tilde{\X}$ is sparse, it is possible to store this matrix and
perform LSI using sparse matrix handling techniques. While this means
that both the \N and \CS methods are unlikely to be of direct value for 
text processing, we choose this example for several reasons.  First, LSI is an
active and evolving field in which singular vectors of the
document-term matrix are used to improve document queries and hence
the application is very relevant.  Second, due to the sparse
structure, we can still compute the eigenvectors of this matrix
and hence compare the approximations directly.  Note that we do not
center $\tilde{\X}$ since this would eliminate the sparsity which
renders computation of the singular vectors possible and would
contrast with the typical 
analysis.  It should be noted, however, that using the \N or
column-sampling methods would enable centering in practice resulting
in an approximation to PCA.

\begin{figure}[h!]
  \centering
  \begin{tabular}{ccc}
 \includegraphics[width=1.7in,trim=25 35 25 50,clip]
 {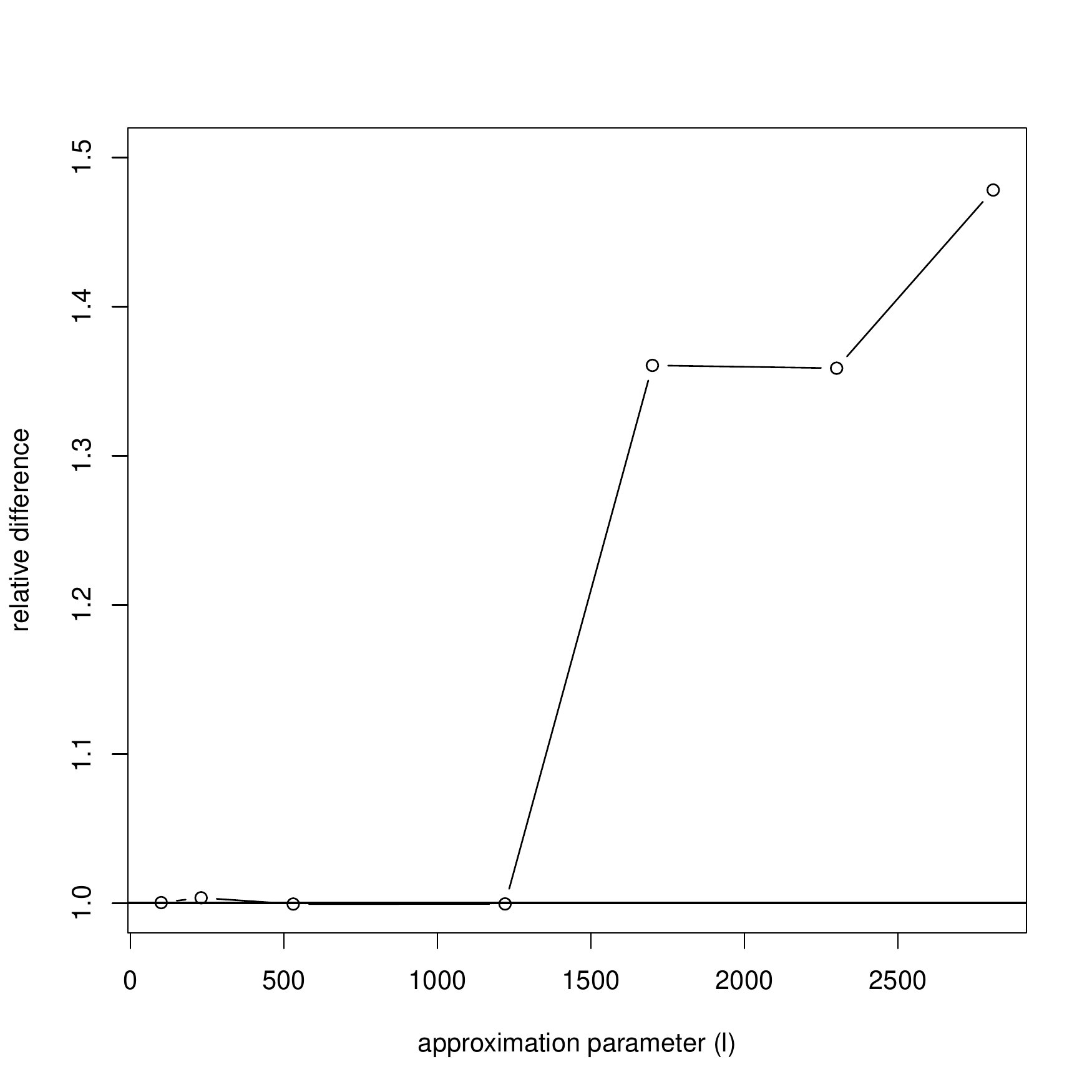}    
 &
 \includegraphics[width=1.7in,trim=25 35 25 50,clip]
 {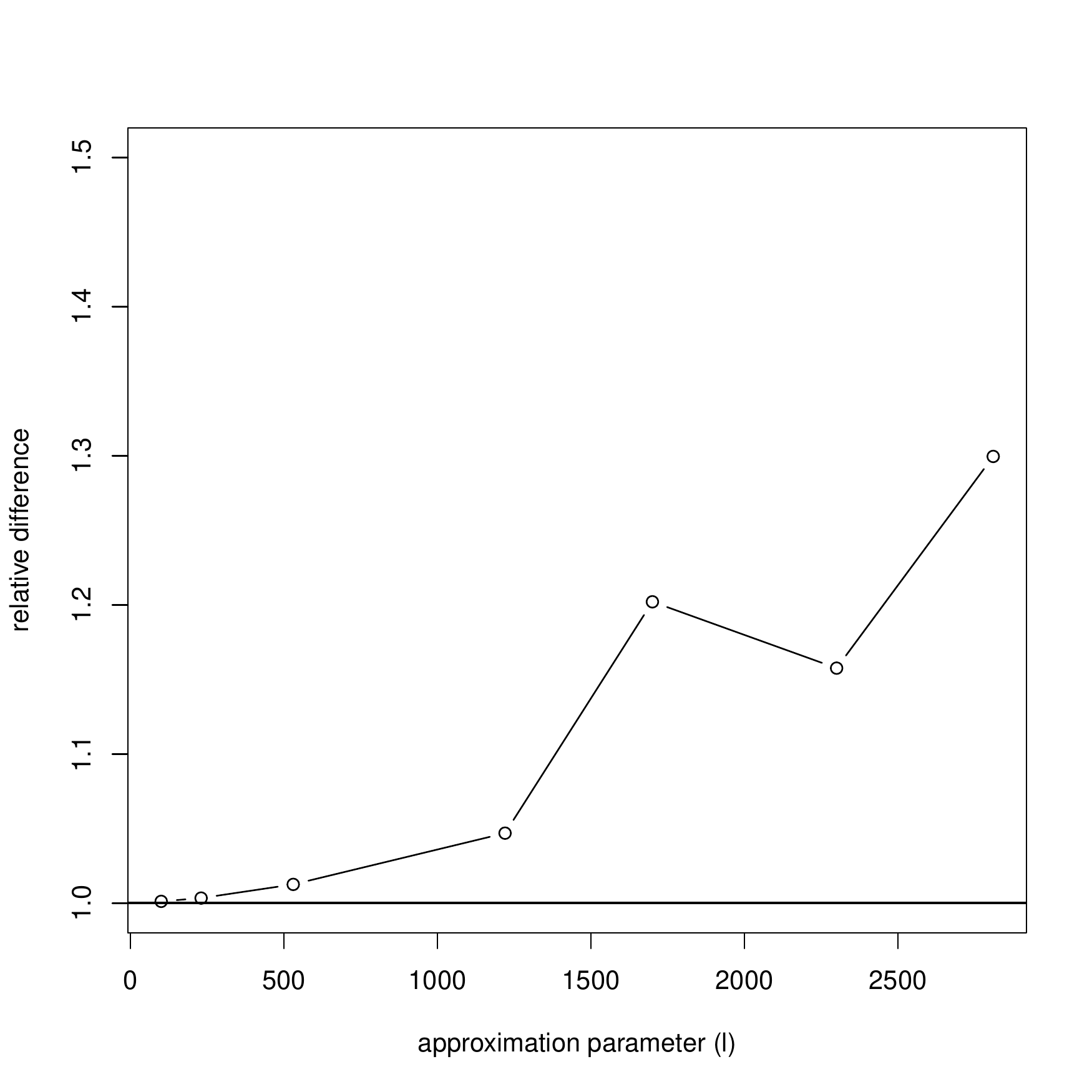}   
  &
  \includegraphics[width=1.7in,trim=25 35 25 50,clip]
 {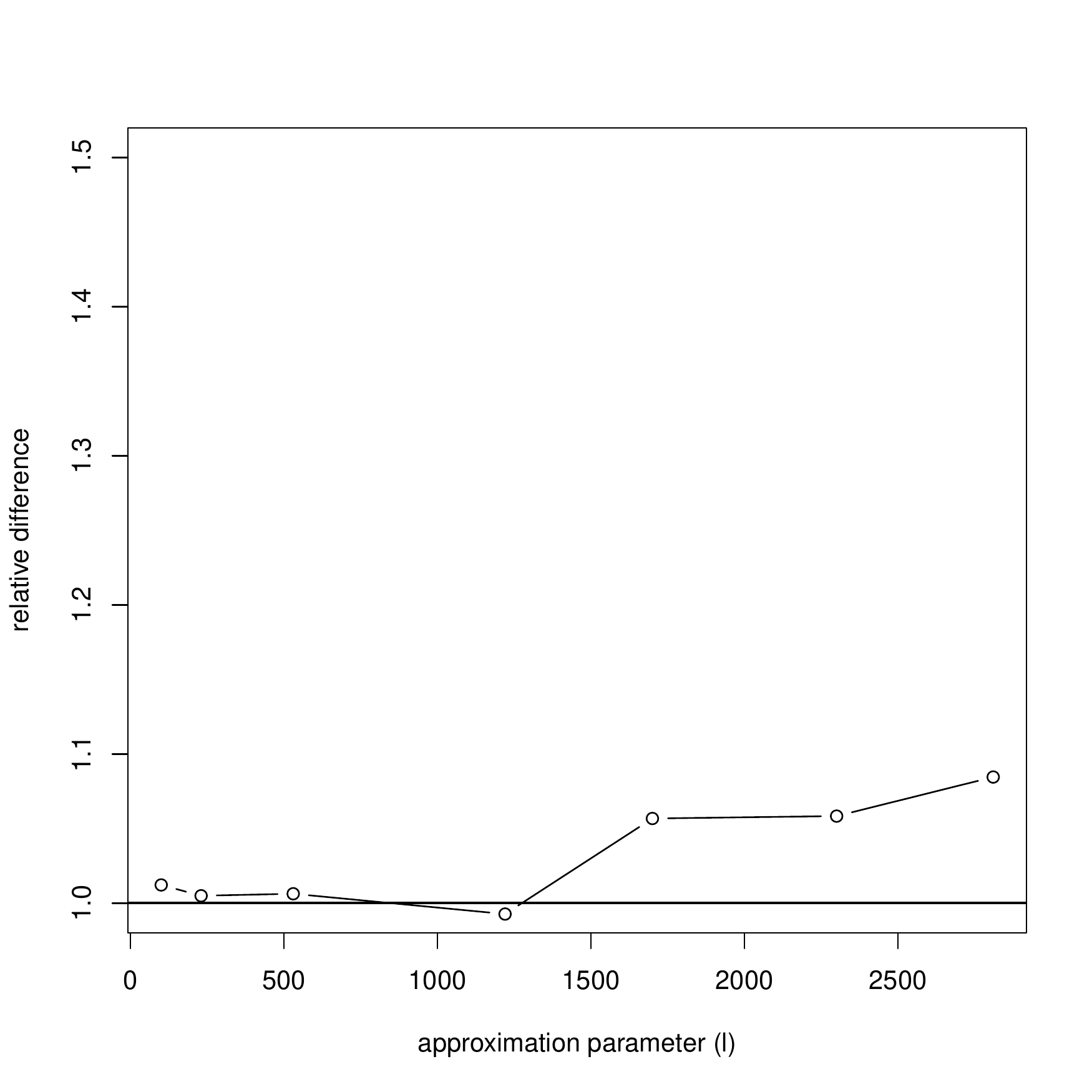}    \\
  ($d = 2$) &   ($d = 3$)  & ($d=50$)   
 \end{tabular}
\caption{The norm difference for the Enron data 
of $V_{nys}$ to $V$ relative to the norm difference
of $V_{cs}$ to $V$ (see equation \eqref{eq:figureMerit}).
The $x$-axis is the grid of $l$ values $\{100,230,530,1220,1700,2300,2810\}$.}
  \label{fig:enronVrel}
\end{figure}

For $d = 2, 3,50$ we compute and plot the relative error given by equation \eqref{eq:figureMerit}
for the grid of $l$ values $\{100,230,530,1220,1700,2300,2810\}$ (see \autoref{fig:enronVrel}). 
Interestingly, for each $d$, smaller values of $l$ have both the \N and \CS method performing
rather similarly.  For larger values of $l$, the \CS method dramatically outperforms the \N
method, although 
this advantage appears to erode for larger values of $d$.  This indicates that in the large $l$ 
regime where the \N method has a decided computational advantage it also performs markedly
worse than the \CS method. This behavior is predicted by the
theoretical results presented in the next section.

\section{Theoretical results}
\label{sec:theory}
With so many possible approximations, it is natural to ask two
questions: (1) what is the cost in approximation accuracy of using the \N
method over the \CS method?  and (2) how do the methods for
approximating $V$ and $U$ compare to each other?  

Comparing eigenvectors to their approximations in the most obvious
fashion can lead to difficulties.
It is not possible to directly measure the distance between $V_d$
and $V_{nys,d}$ or $V_{cs,d}$ because these matrices are not
uniquely defined.  Each eigenvector and its
approximation is only identified up to a sign change.
Also, if 
eigenvalues of $\inner$, $\outer$, $\inner_{11}$, $\outer_{11}$, or
singular values of $L(\inner)$ or $L(\outer)$ are repeated, then the eigenvectors or their
approximations are not uniquely defined. Hence, it is cumbersome to
compare $V_d$ to $V_{nys,d}$ or $V_{cs,d}$ directly via
a matrix norm.  Even comparing the column spaces of the 
matrices is not appropriate as,
for any orthogonal $\O\in \mathbb{R}^{d\times d}$, $\ran(V_d \O)
=\ran(V_d)$. Instead, we compare the {\it
  subspaces} spanned by the eigenvectors $V_d$ (or $U_d$) and the
  relevant
approximations.  This not only provides a coherent metric, but is
exactly the relevant quantity for many principal component based
applications.

For any two subspaces
$\G$ and $\H$ with associated orthogonal projections $\Pi_\G$ and
$\Pi_\H$, we define the distance between $\G$ and $\H$ to be
\begin{equation}
  \Delta(\G,\H) = ||\Pi_\G - \Pi_\H ||_F.
\end{equation}
 We will use $\Delta$ as our loss function for examining how well we
can recover the PCA generated subspaces
given that we are constrained to using an approximation.



%

\textbf{Results.} Define $\V_d := \ran(V_d)$.  Using $V_{nys,d}$ as the \N approximation to $V_d$,
the \N approximation of 
$\V_d$ is
$\V_{nys,d} = \ran( V_{nys,d})$, which has orthogonal projection
$V_{nys,d} ((V_{nys,d})^\top V_{nys,d})^{-1} (V_{nys,d})^{\top}$.
Likewise, the column sampling approximation of the subspace $\V_d$ is
$\V_{cs,d} = \ran( V_{cs,d})$, which has orthogonal projection
$V_{cs,d}{V_{cs,d}}^{\top}$.  

In this section, we provide bounds on the distance between
the subspace $\V$ and its \N and \CS approximations.
As an aside, comparing these two approximations to the population-level
eigenvectors of 
$\mathbb{E}[\tilde{X}_1 \tilde{X}^\top_1] - \mathbb{E}[\tilde{X}_1]\mathbb{E}[\tilde{X}_1]^{\top}$ 
is of interest, as it is possible that $V_{cs}$, for 
example, might be farther from $V$ than $V_{nys}$ but closer to 
$V(\mathbb{E}[\tilde{X}_1 \tilde{X}^\top_1] - \mathbb{E}[\tilde{X}_1]\mathbb{E}[\tilde{X}_1]^{\top})$.  However, the perspective of this paper is
that the data analyst wishes to conduct a principal components
analysis but cannot due to computational or size constraints.
Therefore, we seek upper bounds on the accuracy loss incurred through
a computational approximation.

We write $\X =
[X_1,\ldots,X_n]^{\top} = [x_1,\ldots,x_p]$ and define for two
matrices $\Aone$ and $\Atwo$, $\gap{\Aone,\Atwo}
= 
\lambda_d(\Aone) - \lambda_{d+1}(\Atwo)
$.
Then we have the following upper bounds, pointwise with respect to the
distribution of both  
$\pi$ and the data.  

\begin{theorem}[\N bound]
  Suppose that $\gap{\inner,\inner_{11}}  = \epsilon$.  
  \begin{align*}
    \Delta(\V_d,\V_{nys,d})
    & \leq 
    \frac{\sqrt{2}}{n\epsilon}\left( 2\sum_{j=l+1}^p\sum_{k=1}^l (x_j^{\top}x_k)^2
      + \sum_{j=l+1}^p\sum_{k=l+1}^p (x_j^{\top}x_k)^2\right)^{1/2} \\
    & \qquad \qquad +
    \sqrt{2}\left( \trace{
        \Omega_d^{\top}(I + \Omega_d \Omega_d^{\top})^{-1} \Omega_d
      }\right)^{1/2},
  \end{align*}
  where $\Omega = \inner_{21} V(\inner_{11}) \Lambda(\inner_{11})^{\dagger}$.
  \label{thm:pcaNystrom}
\end{theorem}
\begin{theorem}[Column-sampling bound]
  Suppose $\gap{\inner,L(\inner)}  = \delta$. Then
  \begin{align*}
    \Delta(\V_d,\V_{cs,d})
    & \leq  
    \frac{1}{\delta n}\left( \sum_{j=l+1}^p\sum_{k=1}^l (x_j^{\top}x_k)^2
      + \sum_{j=l+1}^p\sum_{k=l+1}^p (x_j^{\top}x_k)^2\right)^{1/2}
  \end{align*}
  \label{thm:pcaCS}
\end{theorem}

\begin{remark}
In  \autoref{thm:pcaNystrom} and \ref{thm:pcaCS}, as the spectral
gap ($\epsilon$ or $\delta$) gets smaller, the bound becomes worse. 
This is analogous to the necessity of a spectral gap for finding an eigenspace for a given matrix $\A$.  Suppose 
$\mathcal{A}_1$ and $\lambda_d(\A) - \lambda_{d+1}(\A) = c$, are the
$d$-dimensional eigenspace and spectral gap of $\A$, respectively.  
The computation of $\mathcal{A}_1$ becomes unstable as $c
\rightarrow 0$, with $c = 0$ implying $\mathcal{A}_1$ is no longer uniquely
defined.

\end{remark}
If we assume additional structure on $\X$ we can more directly compare
the \N and \CS methods.  Suppose we know, or are willing to impose,
some correlation structure on our data, as given by the following
condition
\begin{condition}
  Define the set $\Xi(r,p) = \{ j,k : j\ne k \textrm{ and } 1 \leq
  j \leq p, r+1 \leq k \leq p\}$.  Then we say that $\X$ has
  $\Xi(r,p)$-coherence $C$ if $\;  \max_{(j,k) \in \Xi(r,p)} x_j^{\top} x_k \leq C$.
  \label{cond:incoherence}
\end{condition}
As $L(\inner)^{\top}L(\inner) = \inner_{11}^2 + \inner_{21}^\top
\inner_{21}$ is the sum of 
nonnegative definite matrices, it must hold that, for any $d$,
$\lambda_d(L(\inner)^{\top}L(\inner)) \geq \lambda_d(\inner_{11}^2)$, which implies that
$\lambda_d(L(\inner)) \geq \lambda_d(\inner_{11})$ as both are nonnegative. This implies the following
inequality and corollary
\begin{equation}
  \frac{1}{\gap{\inner,\inner_{11}}} \leq \frac{1}{\gap{\inner,L(\inner)}}.
\end{equation}

\begin{corollary}
  Suppose $\X$ has $\Xi(l,p)$-coherence $C$ and $\gap{\inner,L(\inner)}  = \delta$.  Then
  \begin{align*}
    \Delta(\V_d,\V_{nys,d}) & \leq C\frac{\sqrt{p^2-l^2}}{n\delta}
    +\sqrt{d- \trace{(V_{nys,d}^\top V_{nys,d})^{-1}}}, \\
    \Delta(\V_d,\V_{cs,d}) & \leq C \frac{\sqrt{(p-l)p}}{n\delta}.
  \end{align*}
  \label{cor:pcaNystromCS}
\end{corollary}

\begin{figure}[t!]
  \centering
  \includegraphics[width=3.25in]{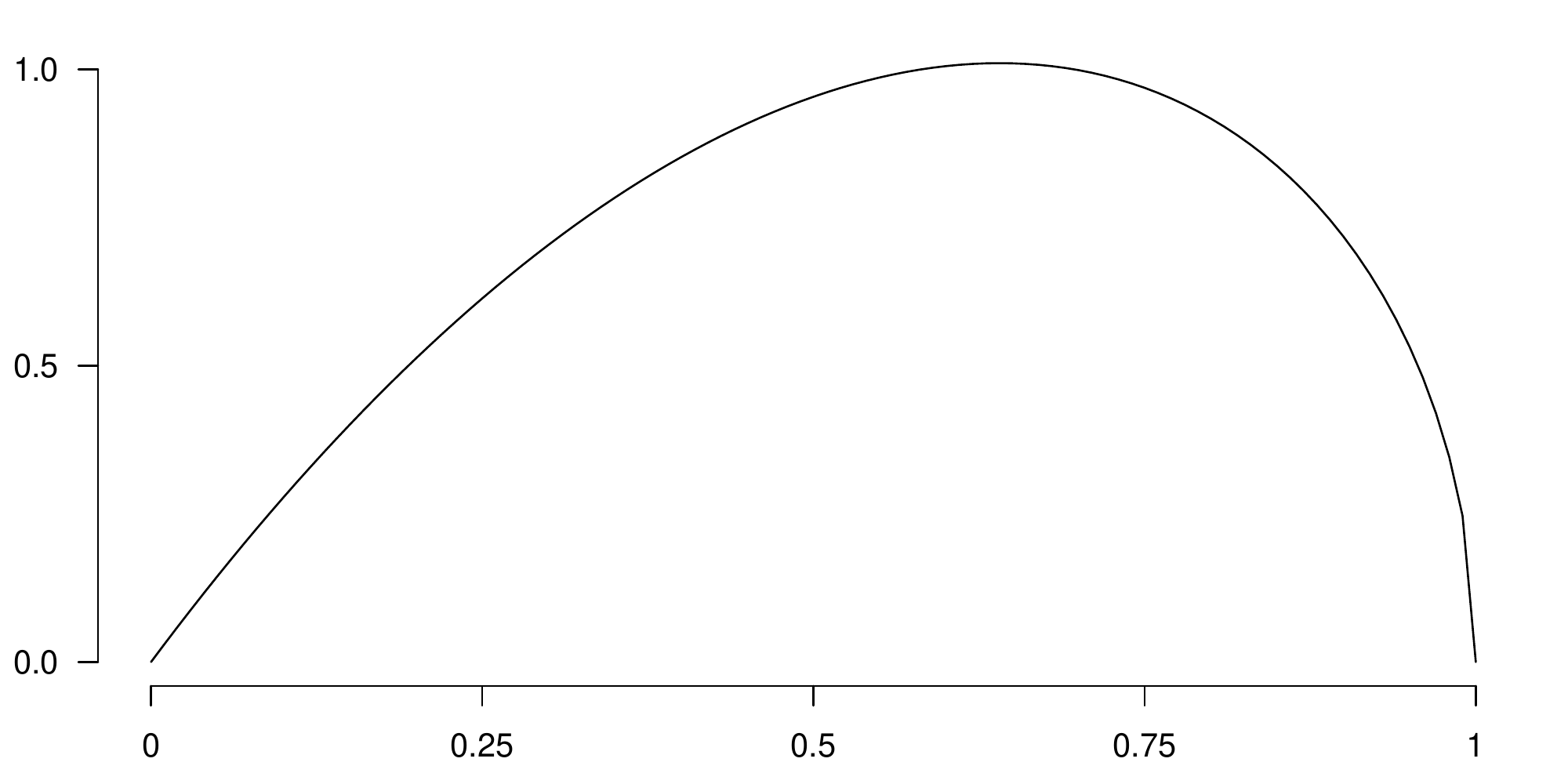}
  \caption{The difference
    $\frac{\sqrt{p^2-l^2}}{n\delta}-\frac{\sqrt{(p-l)p}}{n\delta}$ as
    a function of $l$ for $p$ fixed. The $y$-axis is the size of
    the difference as a percentage of the maximum. The $x$-axis
    is $l/p$.} 
  \label{fig:lpcomp}
\end{figure}

The usefulness of \autoref{cor:pcaNystromCS} is that the upper
bounds are easier to compare.  We pay a penalty
for using \N over \CS that is comprised of two parts.  The first
can be interpreted as the relative behavior due to the choice of $l$.
For fixed $l$, the limit as $p\rightarrow\infty$ is $l/2$, so
the difference between the methods is asymptotically constant. 
Furthermore, in the fixed $p$ scenario in
\autoref{sec:empiricalResults}, the difference is quadratic in
$l$. This behavior is illustrated in \autoref{fig:lpcomp}.
For the second term, any rank $d$ orthogonal matrix $\O$ has
$\textrm{trace}(\O^{\top}\O) = d$.  Therefore, we can interpret the second term in
the \N bound as being a measure of the deviation from orthogonality
that is inherent in the \N method.  Note that the \CS method produces
an orthogonal approximation and has no such second term. This
comparison partially explains the results from the Enron example
discussed in \autoref{sec:enron}. For moderately large $l$ ($10\%$ of $p$ and
$\sim7\%$ of $n$), \CS shows improvement over the \N method, while for smaller $l$
relative to $n$, the difference is negligible.

For approximating $U$, there are many choices.
Not only is there a \N versus \CS method trade-off, we can either use
the approximation via the matrix $\outer$ (that is, $U_{nys}$ or $U_{cs}$), the indirect approximation of
$U$ with the weighted coordinates of $\X$ in the basis found by approximating $V$ (that is, $\hat{U}_{nys}$ or $\hat{U}_{cs}$), or by directly using the orthogonalization of $\x_1$ (that is, $\hat{U}$).
In what follows, we compare the \N versions of these approximations.  

After some manipulations analogous to equations \eqref{eq:Vnys} and \eqref{eq:VnysExpansion}, we find that
\begin{equation}
  U_{nys} 
  = 
  \X V(\X_1) \Lambda(\X_1)^{\dagger}
  \quad \textrm{and} \quad
  \hat{U}_{nys}
  = 
  \X
  \begin{bmatrix}
    V(\x_1) \\ \x_2^{\top} U(\x_1)\Lambda(\x_1)^{\dagger}
  \end{bmatrix}
  \Lambda(\x_1)^{\dagger}. 
  \label{eq:UhatNys}
\end{equation}
If we rewrite equation \eqref{eq:UhatNys} using
\begin{equation}
  \Psi
  :=
  \begin{bmatrix}
    V(\x_1) \\ \x_2^{\top} U(\x_1)\Lambda(\x_1)^{\dagger}
  \end{bmatrix},
\end{equation}
we see that $U_{nys}$ and $\hat{U}_{nys}$ generate subspaces in a related manner:
\begin{equation}
  U_{nys} 
  = 
  \left[\X \frac{v_1(\X_1)}{ \lambda_1(\X_1)}, \ldots, \X \frac{v_l(\X_1)}{ \lambda_l(\X_1)}\right]
  \quad \textrm{and} \quad
  \hat{U}_{nys}
  = 
  \left[\X \frac{\psi_1}{ \lambda_1(\x_1)}, \ldots, \X \frac{\psi_l}{ \lambda_l(\x_1)}\right],
  \end{equation}
where $\psi_j = [v_j(\x_1)^{\top},
(\x_2^{\top}u_j(\x_1)/\lambda_j(\x_1))^{\top} ]^{\top}$ is the
$j^{th}$ column of $\Psi$. Therefore, these methods are both special
cases of Galerkin methods for discretizing an operator in an integral 
equation.  This realization suggests an interesting extension of these methods using different Galerkin bases
that warrants further investigation.

Further simplifications can be made:
\begin{align}
   U_{nys}
   &=
  \begin{bmatrix}  
    U(\X_1) \\ \X_2^{\top} V(\X_1)\Lambda(\X_1)^{\dagger}
  \end{bmatrix} \\
  \intertext{and}  
  \hat{U}_{nys} &= U(\x_1) + \x_2\x_2^{\top} U(\x_1)  \Lambda(S_{11})^{\dagger}.
\end{align}
Remembering that $\hat{U} = U(\x_1)$, $\hat{U}_{nys}$ can be seen to
be a perturbation of $\hat{U}$ by the matrix $\x_2\x_2^{\top} U(\x_1)  \Lambda(S_{11})^{\dagger}$. 
Supposing that $\x_1$ and $\x_2$ are orthogonal to each other, then for any
vector $x \in \mathbb{R}^l$, $||\hat{U}x||_2 \leq ||\hat{U}_{nys}x||_2$ and hence $\hat{U}_{nys}$ includes more
range space information than $\hat{U}$ 
by preserving the part of vectors in $\ran(\x_2)$.  Also, $U_{nys}$ needs to be
``extended'' to $\mathbb{R}^p$ by concatenating $U(\X_1)$ with $\X_2^{\top} V(\X_1)\Lambda(\X_1)^{\dagger}$
while $\hat{U}_{nys}$ is already in the ``correct'' space.  

These facts all point to $\hat{U}_{nys}$ being the better
of the three methods.  Indeed, the results from \autoref{sec:empiricalResults} provide additional
evidence for this conclusion.

\section{Discussion}
\label{sec:discussion}
For very large problems, PCA
 requires addressing computational and memory
constraints. The computational complexity of
PCA is dominated by finding
the SVD of $\X$.  Hence, some approximations are required to
accomplish a PCA-based reduction of a very large data set.  In this
paper, we investigate the \N and \CS methods for approximating
the eigenvectors  of large, dense matrices.

While the results we present are novel and useful,
 there are a number of potential avenues for further research.
A comparison of the subspaces generated by more general matrix sketches
\citep{halko2011finding,tropp2011improved} would be
of interest and could provide better guidance for practitioners.
Also, in our analysis of the \N and \CS methods, we ignore the question of
how to select the columns of $\X$, sampling them uniformly. However,
other results suggest non-uniform sampling will yield better
approximations to $\X$.
Choices of sampling methods or the use of different sketching matrices
represent additional areas for future work.  

Lastly, 
the centering of the matrix $\X$ can be accomplished in a massively parallel fashion 
using a distributed computing approach such as Map-Reduce.  If the
vector of column means and the associated number of observations is saved, then 
the column means can be readily updated if a new observation is recorded.  However, 
recentering would require another pass through each column of the matrix and hence would be quite slow. Adapting
these approaches to streaming data requires further research.

\section*{Acknowledgements}

We thank the editor, associate editor, and one referee for their
helpful comments and suggestions. Darren Homrighausen is partially
supported by NSF Grant DMS--14-07543. Daniel J.\ McDonald is partially
supported by NSF Grant DMS--14-07439.

\bibliographystyle{ECA_jasa}
\bibliography{../jcgs/references}

\end{document}